\documentclass[a4paper]{article}
\usepackage[a4paper]{geometry}

\usepackage{natbib}
\setcitestyle{authoryear}

\usepackage{pifont,verbatim,anyfontsize}

\usepackage[utf8]{inputenc}
\usepackage[T1]{fontenc}
\usepackage[hyperfootnotes=false,citecolor=black]{hyperref}
\usepackage{url}
\usepackage{booktabs}
\usepackage{nicefrac}
\usepackage{microtype}

\usepackage{array,arydshln,floatrow}

\usepackage[justification=justified]{caption}
\usepackage{subcaption}
\usepackage{graphicx}
\graphicspath{{figures/}}
\usepackage{ifpdf}
\ifpdf
\fi

\usepackage{amsmath,amsthm,amsfonts,amssymb,bm,mathtools}

\renewcommand{\tanh}{\mathtt{tanh}}
\newcommand{\sigm}{\mathtt{sigm}}
\newcommand{\relu}{\mathtt{relu}}
\newcommand{\elu}{\mathtt{elu}}
\newcommand{\sech}{\mathtt{sech}}
\newcommand{\trans}{\operatornamewithlimits{trans}}
\newcommand{\diag}{\mathtt{diag}}

\def\dTheta{\mathrm{d}\Theta}
\def\dOmega{\mathrm{d}\Omega}

\usepackage{cleveref}
\crefname{subsection}{Subsec.}{Subsecs.}
\crefname{section}{Sec.}{Secs.}

\newtheorem{theorem}{Theorem}

\newtheorem{proposition}[theorem]{Proposition}
\newtheorem{definition}[theorem]{Definition}

\newcommand{\defeq}{\stackrel{def}{=}}

\title{Relative Natural Gradient for Learning Large Complex Models}

\author{Ke Sun\\
\'Ecole Polytechnique\\
\texttt{sunk.edu@gmail.com}
\and
Frank Nielsen\\
\'Ecole Polytechnique\\
Sony CSL\\
\texttt{Frank.Nielsen@acm.org}}

\begin{document}

\maketitle

\begin{abstract}
Fisher information and natural gradient provided deep insights and powerful
tools to artificial neural networks. However related analysis becomes more and
more difficult as the learner's structure turns large and complex. This paper
makes a preliminary step towards a new direction. We extract a local component
of a large neuron system, and defines its relative Fisher information metric
that describes accurately this small component, and is invariant to the other
parts of the system. This concept is important because the geometry structure is
much simplified and it can be easily applied to guide the learning of neural
networks. We provide an analysis on a list of commonly used components, and
demonstrate how to use this concept to further improve optimization.
\end{abstract}

\section{Introduction}

\sloppy



The Fisher Information Metric (FIM) $\mathcal{I}(\bm\Theta)=(\mathcal{I}_{ij})$
of a statistical parametric model $p(\bm{x}\,\vert\,\bm\Theta)$ of order $D$ is
defined by a $D\times D$ positive semidefinite (psd) matrix
($\mathcal{I}(\bm\Theta) \succeq 0$) with coefficients 
\begin{equation}\label{eq:fim}
\mathcal{I}_{ij}
= E_p\left[
\frac{\partial l}{\partial\Theta_i} \frac{\partial l}{\partial\Theta_j}
\right],
\end{equation}
where $l(\bm\Theta)$ denotes the log-likelihood function $\ln{p}(\bm{x}\,\vert\,\bm{\Theta})$.
Under light regularity conditions, \Cref{eq:fim} can be rewritten equivalently as
\begin{equation}\label{eq:fim2}
\mathcal{I}_{ij}
=
-E_p\left[\frac{\partial^2 l}{\partial\Theta_i\partial\Theta_j}\right]
=
4\int
\frac{\partial \sqrt{p(\bm{x}\,\vert\,\bm{\Theta})}}{\partial\Theta_i}
\frac{\partial\sqrt{p(\bm{x}\,\vert\,\bm\Theta)}}{\partial\Theta_j}
\mathrm{d}\bm{x}.
\end{equation}
For regular natural exponential
families (NEFs) $l(\bm\Theta) = \bm\Theta^\intercal \bm{t}(\bm{x}) -
F(\bm\Theta)$ (log-linear models with sufficient statistics $\bm{t}(\bm{x})$),
the FIM is $\mathcal{I}(\bm\Theta)=\nabla^2 F(\bm\Theta)\succ 0$, the Hessian of the
moment generating function (mgf). Although exponential families can
approximate arbitrarily {\em any} smooth density~\citep{cobb83}, the mgf may
not be available in closed-form nor computationally tractable~\citep{montanari15}.  
Besides the fact that learning machines usually have
often singularities~\citep{SLT-2009} ($\vert{}\mathcal{I}(\bm\Theta)\vert=0$, not full rank)
characterized by plateaux in gradient learning, computing/estimating the FIM of
a large learning system is very challenging due to the {\em finiteness of data}, and
the {\em large number} $\frac{D(D+1)}{2}$ of matrix coefficients to evaluate.
Moreover, gradient descent techniques require to invert this large matrix and
to tune the learning rate. 
The FIM is \emph{not} invariant and depends on the
parameterization: $\mathcal{I}_{\bm\Theta}(\bm\Theta)= \bm{J}^\intercal
\mathcal{I}_{\bm{\Lambda}}(\bm\Lambda) \bm{J}$ where $\bm{J}$ is the Jacobian matrix
$J_{ij}=\frac{\partial \Lambda_i}{\partial \Theta_j}$.
Therefore one may ponder whether we can always find a suitable parameterization that yields a diagonal FIM that is straightforward to invert.
This fundamental problem of parameter orthogonalization was first investigated by Jeyffreys~\citeyearpar{Jeffreys-1939} for decorrelating the estimation 
of the {\em parameters of interest} from the {\em nuisance parameters}.
Fisher diagonalization yields parameter
orthogonalization~\citep{cox87}, and prove useful when estimating
$\hat{\bm\Theta}$ using MLE that is asymptotically normally distributed,
$\hat{\bm\Theta}_n=G(\bm\Theta,\mathcal{I}^{-1}(\bm\Theta)/\sqrt{n})$, where
$G(\bm\theta_1, \bm\theta_2)$ denotes a univariate or multivariate Gaussian
distribution with mean $\bm\theta_1$ and variance $\bm\theta_2$, and efficient
since the variance of the estimator matches the Cram\'er-Rao lower bound.
Using the chain rule of differentiation of calculus, this amounts to find a suitable parameterization  $\Omega=\Omega(\Theta)$  satisfying

$$
\sum_{i,j} E\left[\frac{\partial^2}{\partial\Theta_i\partial\Theta_j}\partial l(x;\theta)\right] \frac{\partial\Theta_i}{\partial\Omega_k}\frac{\partial\Theta_j}{\partial\Omega_l}  =0,\quad \forall k\not = l.
$$

Thus in general, we end up with $\binom{D}{2}=\frac{D(D-1)}{2}$ (non-linear) partial  differential equations to satisfy~\citep{OrthogonalParameters-1950}.
Therefore, in general there is no   solution when $\binom{D}{2}>D$, that is when $D>3$.
When $D=2$, the single differential equations is  usually solvable and tractable, and the solution may not be unique:
For example, Huzurbazar~\citeyearpar{OrthogonalParameters-1950} reports two orthogonalization schemes for  the  location-scale families $\{\frac{1}{\sigma}p_0(\frac{x-\mu}{\sigma})\}$ that include the Gaussian family and the Cauchy family.
Sometimes, the structure of the differential equation system yields a solution: For example,  Jeffreys~\citeyearpar{Jeffreys-1939} reported a parameter orthogonalization for Pearson's distributions of type I which is of order $D=4$.
Cox and Reid~\citeyearpar{cox87} further investigate this topic with application to conditional inference, and provide examples (including the Weibull distribution).

From the viewpoint of geometry, the FIM induces a Riemannian manifold with metric
tensor $g(\bm\Theta)=\mathcal{I}(\bm\Theta)$.
When the FIM may be degenerate, this yields a pseudo-Riemannian manifold~\citep{GeNGA-2014}. 
In differential geometry, orthogonalization amounts to transform the square length infinitesimal element 
$g_{ij}\dTheta_i\Theta_j$ of a Riemannian geometry into an orthogonal system $\omega$ with  
 matching square length infinitesimal element  $\Omega_{ii}\dOmega_i\dOmega_j$.
However, such a global orthogonal metric does not exist~\citep{OrthogonalParameters-1950} when $D>3$ for an arbitrary metric tensor although interesting Riemannian parameterization structures may be derived in Riemannian 4D geometry~\citep{BlockDiagonal4D-2009}.
For NEFs, the FIM can be made {\em block-diagonal} easily by
using the {\em mixed coordinate system}~\citep{amari16} $(\bm\Theta_{1:k},\bm{H}_{k+1:D})$,
where $\bm{H}=E_p[\bm{t}(\bm{x})]=\nabla F(\bm\Theta)$ is the moment parameter,
for any $k\in\{1, ..., D-1\}$, where $\bm{v}_{[b:e]}$ denotes the subvector $(v_b,
..., v_e)^\intercal$ of $\bm{v}$.  
The geometry of NEFs is a dually flat structure~\citep{amari16} induced by the convex mgf, the
potential function.  It defines a dual affine coordinate systems 
$e^i=\partial_i=\frac{\partial}{\partial H_i}$ and 
$e_j=\partial^j=\frac{\partial}{\partial\Theta^j}$ that are orthogonal: 
$\langle{e^i},{e_j}\rangle=\delta_j^i$,
where $\delta_j^i=1$ iff $i=j$ and $\delta_j^i=0$ otherwise.
Those dual affine coordinate systems are defined up to an affine 
invertible transformation $\bm{A}$:
$\tilde{\bm\Theta} = \bm{A}\bm\Theta + \bm{b}$, $\tilde{\bm H} =
\bm{A}^{-1}\bm H+\bm{c}$, where $\bm{b}$ and $\bm{c}$ are constants.
In particular, for any order-2 NEF ($D=2$), we can {\em always} obtain two mixed
parameterizations $(\Theta_1,{H}_2)$ or $({H}_1,\Theta_2)$. 

The FIM $g(\bm\Theta)$ or $\mathcal{I}(\bm\Theta)$ by definition is an
expectation. If its Hessian form in \cref{eq:fim2} is computed based on
a set of empirical observations $\{\bm{x}_k\}$, as
\begin{equation}
\bar{g}(\bm\Theta)
=
\overline{ -\frac{\partial^2 l_k}{\partial\Theta_i\partial\Theta_j} },
\end{equation}
where ``$\overline{~\cdot~}$'' denotes the sample average over $\{\bm{x}_k\}$,
the resulting metric is called the \emph{observed FIM}~\citep{efron78}. It is
useful when the underlying distribution is not available. When the number of
observations increases, the observed FIM becomes more and more close to the FIM.

 
Past works on FIM-based approaches mainly focus on how to approximate the global
FIM into a block diagonal form using the gradient of the cost
function~\citep{roux08,martens10,pascanu14,martens15}. This global approach
faces the analytical complexity of learning systems. The approximation error
increases as the system scales up and as complex and dynamic structures emerge.

This work aims at a different \emph{local} approach. The idea is to {\em accurately}
describe the information geometry in a local subsystem of the big learning
system, which is {\em invariant} to the scaling up and structural change of the
global system, so that the local machinery, including optimization, can be
discussed regardless of the other parts.

For this purpose, a novel concept, {\em Relative Fisher Information Metric} (RFIM), is defined. Unlike
the traditional geometric view of a high-dimensional parameter manifold, RFIMs
defines {\em multiple projected low-dimensional geometry of subsystems}. This geometry
is correlated to the parameters beyond the subsystem and is therefore considered {\em dynamic}.
It can be used to characterize the efficiency of a local learning process.  
Taking this stance
has potentials in deep learning because a big learning system can be decomposed
into many local components, i.e. layers. This paper will make clear the concept
of RFIM, provide proof-of-concept experiments, and discuss its theoretical advantages.

The paper is organized as follows. \Cref{sec:review} reviews natural gradient
within the context of Multi-Layer Perceptrons (MLP). \Cref{sec:rfim} presents
the concept of RFIM, and gives detailed formulations of several commonly used
subsystems. \Cref{sec:norm} discusses the advantages of using RFIM as compared
to FIM. \Cref{sec:rng} shows how to use the RFIMs given by \cref{sec:rfim} to
optimize neural networks, with an algorithm framework and several proof-of-concept
experiments. \Cref{sec:con} concludes this work and further hint at perspectives.

\section{Natural Gradient of Neural Networks}\label{sec:review}

\begin{figure}[t]
\includegraphics[width=.35\textwidth]{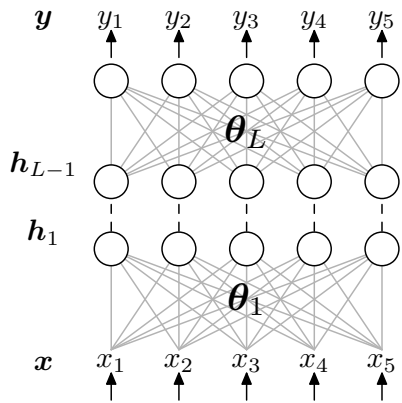}
\caption{A multilayer perceptron (MLP): A feedforward hierarchical multi-layer machine with hidden layers. The parameter vector $\Theta=(\theta_1, \ldots, \theta_L)$ stores the connection weights.}\label{fig:mlp}
\end{figure}

Consider a MLP as depicted in \cref{fig:mlp},
whose statistical model is the following {\em conditional distribution}
\begin{equation}
p(\bm{y}\,\vert\,\bm{x},\bm\Theta)
=
\sum_{\bm{h}_1,\cdots,\bm{h}_{L-1}}
p(\bm{y}\,\vert\,\bm{h}_{L-1},\bm\theta_{L})
\cdots{}
p(\bm{h}_2\,\vert\,\bm{h}_1,\bm\theta_2)
p(\bm{h}_1\,\vert\,\bm{x},\bm\theta_1),
\end{equation}
where the often intractable sum over $\bm{h}_1,\cdots,\bm{h}_{L-1}$ can be get
rid off by deteriorating $p(\bm{h}_1\,\vert\,\bm{x},\bm\theta_1)$, $\cdots$,
$p(\bm{h}_{L-1}\,\vert\,\bm{h}_{L-2},\bm\theta_{L-1})$ to Dirac's deltas $\delta$, and let
merely the last layer $p(\bm{y}\,\vert\,\bm{h}_{L-1},\bm\theta_{L})$ be stochastic.
Note that Restricted Boltzmann Machines~\citep{nair10,montavon12} (RBMs), and dropout~\citep{wager13} do consider $\bm{h}$ to be stochastic.

The tensor metric of the neuromanifold~\citep{amari95} $\mathcal{M}_{\bm\Theta}$,
consisting of all MLPs with the same architecture but different parameter
values, is locally defined by the FIM. Because that a MLP corresponds to a
conditional distribution, its FIM by \cref{eq:fim} is a function of the input
$\bm{x}$. By taking an empirical average over the input samples $\{\bm{x}_i\}$,
the FIM of a MLP has the following expression  
\begin{equation}
g( \bm{\Theta} )
=
\frac{1}{n}\sum_{i=1}^n
E_{p(\bm{y}\,\vert\,\bm{x}_i, \bm\Theta)}
\left[
\frac{\partial l_i }{\partial\bm\Theta}
\frac{\partial l_i }{\partial\bm\Theta^\intercal}
\right]
=
- \frac{1}{n}\sum_{i=1}^n
E_{p(\bm{y}\,\vert\,\bm{x}_i, \bm\Theta)}
\left[
\frac{\partial^2 l_i}{\partial\bm\Theta \partial\bm\Theta^\intercal}
\right],
\end{equation}
where $l_i(\bm\Theta)=\ln p(\bm{y}\,\vert\,\bm{x}_i,\,\bm\Theta)$
denotes the conditional log-likelihood function wrt $\bm{x}_i$.

Just like a Mahalanobis metric, $g(\bm{\Theta})$ can be used to measure the
distance between two neural networks locally around
$\bm\Theta\in\mathcal{M}_{\bm\Theta}$. A learning step makes a tiny movement $\delta\bm\Theta$
on $\mathcal{M}_{\bm\Theta}$ from $\bm\Theta$ to $\bm\Theta+\delta\bm\Theta$.
According to the FIM, the infinitesimal square distance
\begin{equation}\label{eq:norm2}
\langle \delta\bm\Theta, \delta\bm\Theta \rangle_{g(\bm\Theta)}
=
\delta\bm\Theta^\intercal
g(\bm{\Theta})
\delta\bm\Theta
=
\frac{1}{n} \sum_{i=1}^n
E_{p(\bm{y}\,\vert\,\bm{x}_i,\,\bm\Theta)}
\left[
\delta\bm\Theta^\intercal
\frac{\partial l_i }{\partial\bm\Theta}
\right]^2
\end{equation}
measures how much $\delta\bm\Theta$ (with a radius constraint) is statistically
along $\frac{\partial l}{\partial\bm\Theta}$, or equivalently how much $\delta\bm\Theta$
affects intrinsically the conditional distribution
$p(\bm{y}\,\vert\,\bm{x},\,\bm\Theta)$.


Consider the negative log-likelihood function $L(\bm\Theta) = -\sum_{i} \ln
p(\bm{y}_i\,\vert\,\bm{x}_i,\bm\Theta)$ wrt the observed pairs
$\{(\bm{x}_i,\bm{y}_i)\}$, we try to minimize the loss while maintaining a small
\emph{cost}, measured geometrically by the square distance
$\langle\delta\bm\Theta, \delta\bm\Theta\rangle_{g(\bm\Theta)}$ on
$\mathcal{M}_{\bm\Theta}$. At $\bm\Theta_t\in\mathcal{M}_{\bm\Theta}$, the
target is to minimize wrt $\delta\bm\Theta$
\begin{equation}\label{eq:lstep}
L(\bm\Theta_t+\delta\bm\Theta) +
\frac{1}{2\gamma}\langle\delta\bm\Theta,\delta\bm\Theta\rangle_{g(\bm\Theta_t)}
\approx
L(\bm\Theta_t) + \delta\bm\Theta^\intercal \bigtriangledown{L}(\bm\Theta_t) +
\frac{1}{2\gamma} \delta\bm\Theta^\intercal g(\bm{\Theta}_t) \delta\bm\Theta,
\end{equation}
where $\gamma>0$ is a learning rate. The optimal solution of the above
\cref{eq:lstep} gives a learning step 
$$\delta\bm\Theta_t = -\gamma
g^{-1}(\bm{\Theta}_t) \bigtriangledown{L}(\bm\Theta_t).
$$
  In this update
procedure, the term $g^{-1}(\bm{\Theta}_t) \bigtriangledown{L}(\bm\Theta_t)$
replaces the role of the usual gradient $\bigtriangledown{L}(\bm\Theta_t)$ and
is called the \emph{natural gradient}~\citep{amari96}.

Although the FIM depends on the chosen parameterization, the natural gradient is \emph{invariant} to reparametrization. Let $\bm\Lambda$ be
another coordinate system and $\bm{J}$ be the Jacobian matrix
$\bm\Theta\to\bm\Lambda$. Then we have 
\begin{equation}\label{eq:pushforward}
g^{-1}(\bm\Theta) \bigtriangledown{L}(\bm\Theta)
=
\left( \bm{J}^\intercal g(\bm\Lambda) \bm{J} \right)^{-1}
\bm{J}^\intercal \bigtriangledown{L}(\bm\Lambda)
=
\bm{J}^{-1} g^{-1}(\bm\Lambda) \bigtriangledown{L}(\bm\Lambda).
\end{equation}
The left-hand-side and right-hand-side of \cref{eq:pushforward} correspond to
exactly the same infinitesimal movement along $\mathcal{M}_{\bm\Theta}$.
However, as the learning rate $\gamma$ is not infinitesimal in practice, natural gradient descent
actually depends on the coordinate system in practice. Other intriguing properties of
natural gradient optimization lie in being free from getting trapped in plateaus
of the error surface, and attaining Fisher efficiency in online learning (see
Sec.  4~\citep{amari98}).

For sake of simplicity, we omit to discuss the case when the FIM is singular.
That is, the set of parameters $\Theta_s$ with zero metric.
This set of parameters forms an analytic variety~\citep{SLT-2009}, and
technically the MLP as a statistical model is said non-regular (and the
parameter $\Theta$ is not identifiable).  The natural gradient has been
extended~\citep{GeNGA-2014} to cope with singular FIMs having positive
semi-definite matrices by taking the Moore-Penrose pseudo-inverse (that
coincides with the inverse matrix for full rank matrices).

In the family of $2^{\mathtt{nd}}$-order optimization methods, a line can be
drawn from natural gradient from the Newton methods, e.g.~\citep{martens10}, by
checking whether the computation of the Hessian term depends on the cost
function. Taking a MLP as an example, the computation of the FIM does not need the
given $\{\bm{y}_i\}$, but averages over all possible $\bm{y}$ generated
according to $\{\bm{x}_i\}$ and the current model $\bm\Theta$. One advantage
of natural gradient is that the FIM is guaranteed to be psd while the Hessian may not.
We refer the reader to related references~\citep{amari00a} for more details.

Bonnabel~\citep{RieSGD-2013} proposed to use the Riemannian exponential map to define a step gradient descent, thus ensuring
to stay on the manifold for any chosen learning rate. Convergence is proven for Hadamard manifolds (of negative curvatures).
However, it is not mathematically tractable to express the exponential map of hierarchical model manifolds.




\section{Relative Fisher Information Metric of Subsystems}\label{sec:rfim}

In general, for large parametric matrices, it is impossible to diagonalize or
decorrelate all the parameters as mentioned above, so that we split instead all random variables
in the learning system into three parts $\bm\theta_f$, $\bm\theta$, $\bm{h}$. The
\emph{reference}, $\bm\theta_f$, consists of the majority of the random variables that
are considered fixed. $\bm\theta$ is the internal parameters of a subsystem wrt
its structure. The \emph{response} $\bm{h}$ is the interface of this subsystem
to the rest of the learning system. This $\bm{h}$ usually carries sample-wise
information to be summarized into the internal parameters $\bm\theta$, so it is
like ``pseudo-observations'', or hidden variables, to the subsystem. Once
$\bm\theta_f$ is given, the subsystem can characterized by the conditional
distribution $p(\bm{h}\,\vert\,\bm\theta,\bm\theta_f)$.  We made the following
definition.

\begin{definition}[RFIM]\label{def:rfim}
Given $\bm\theta_f$, the RFIM~\footnote{We use the same term ``relative
FIM''~\citep{zegers15} with a different definition.} of $\bm\theta$
wrt $\bm{h}$ is
\begin{equation}\label{eq:rfim}
g^{\bm{h}}\left( \bm\theta\,\vert\,\bm\theta_f \right)
\defeq
E_{p(\bm{h}\,\vert\,\bm\theta,\,\bm\theta_f)}
\left[
\frac{\partial}{\partial\bm\theta}
\ln{p}(\bm{h}\,\vert\,\bm\theta,\,\bm\theta_f)
\frac{\partial}{\partial\bm\theta^\intercal}
\ln{p}(\bm{h}\,\vert\,\bm\theta,\,\bm\theta_f)
\right],
\end{equation}
or simply $g^{\bm{h}}\left( \bm\theta \right)$, corresponding to the estimation
of $\bm\theta$ based on observations of $\bm{h}$ given $\bm\theta_f$.
\end{definition}
When we choose $\bm{h}$ to be the observables,  usually denoted by $\bm{x}$, and
choose $\bm\theta$ to be all free parameters $\bm\Theta$ in the learning system,
then RFIM becomes FIM: $g(\bm\Theta)=g^{\bm{x}}(\bm\Theta)$. What is novel is
that \emph{we can choose the response $\bm{h}$ to be other than the raw observables
to compute Fisher informations of subsystems}, specially dynamically during the
learning of machines. To see the meaning of RFIM, similar to \cref{eq:norm2},
the infinitesimal square distance
\begin{equation}
\langle\delta\bm\theta,\delta\bm\theta\rangle_{g^{\bm{h}}(\bm\theta)}
=
E_{p(\bm{h}\,\vert\,\bm\theta,\,\bm\theta_f)}
\left[
\delta\bm\theta^\intercal
\frac{\partial}{\partial\bm\theta}
\ln{p}(\bm{h}\,\vert\,\bm\theta,\,\bm\theta_f)
\right]^2
\end{equation}
measures how much $\delta\bm\theta$ impacts intrinsically the conditional
distribution featuring the subsystem. We also have the following proposition, following 
straightforwardly from \cref{def:rfim}.
\begin{proposition}
If $\bm\theta_1$ consists of a subset of $\bm\theta_2$ so that
$\bm\theta_2 = (\bm\theta_1, \tilde{\bm\theta}_1)$, then
$\mathcal{M}_{\bm\theta_1}$ with the metric $g^{\bm{h}}\left(
\bm\theta_1\right)$ has the same Riemannian geometry with a sub-manifold of
$\mathcal{M}_{\bm\theta_2}$ with the metric $g^{\bm{h}}\left(
\bm\theta_2\right)$, when $\tilde{\bm\theta}_1$ is given.
\end{proposition}
When the response $\bm{h}$ is chosen, then different splits of
$(\bm\theta,\bm\theta_f)$ correspond to the same ambient geometry. In fact, the
particular case of a mixed coordinate system (that is not an affine coordinate
system) induces in information geometry~\citep{amari16} a dual pair of orthogonal
e- and m- orthogonal foliations. Our splits in RFIMs consider non-orthogonal
foliations that provide the factorization decompositions of the whole manifold
into submanifolds, that are the leaves of the foliation~\citep{amari00}.

\begin{figure}[t!]
\centering
\includegraphics[width=\textwidth]{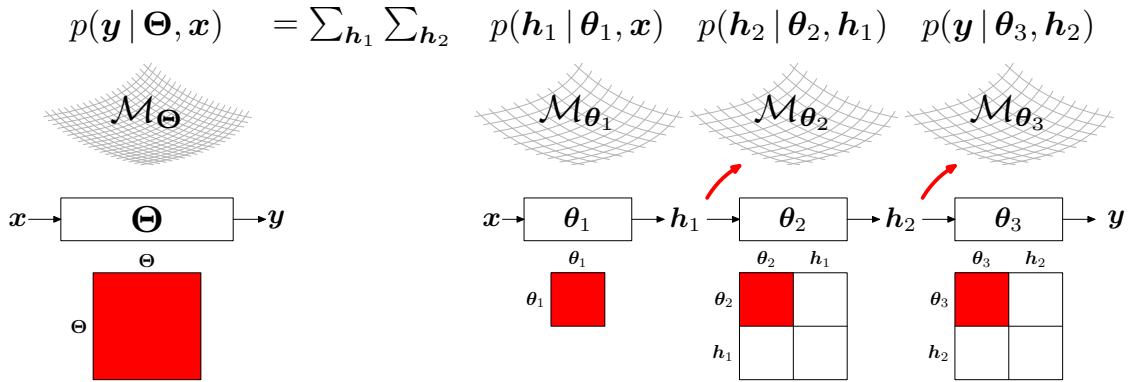}
\centering
\caption{(left) The traditional global geometry of a MLP with two hidden layers;
(right) information geometry of subsystems. The red arrows mean that the pointed
geometry is \emph{dynamic} wrt to the variable, i.e., the value of the variable
affects the geometry. The square under the (sub-)system means the (R-)FIM is
computed by (i) computing the FIM in the traditional way wrt all free parameters
that affect the system output; (ii) choosing a sub-block as shown by the red
square that contains only the internal parameters of the (sub-)system.
The rest variables are regarded as the reference.
}\label{fig:geometry} \end{figure}

\Cref{fig:geometry} shows the traditional global geometry of a learning system,
as compared to the information geometry of subsystems defined by RFIMs.  The red
arrows means that the pointed geometry structure is dynamic and varies with the
reference variable. In MLPs, the subsystems, i.e. layers, are supervised. Their
reference $\bm\theta_f$ also carries sample-wise information.

One should not confuse RFIM with the diagonal blocks of FIM. Both their meaning
and expression are different. RFIM is computed by {\em integrating out} hidden
variables, or the output of subsystems. FIM is always computed by integrating
out the observables. 

In the following we analyze accurately several commonly used RFIMs. Note that,
the FIMs of small parametric structures such as single neurons have been studied
for a long time~\citep{amari96}. Although with similar expressions, we are
looking at a component embedded in a large system rather than a small single
component system. These are two different concepts, and only the former can be
used to guide large learning systems.

\subsection{RFIMs of One Neuron}\label{subsec:1neuron}

We start from the RFIM of single neuron models. 

\subsubsection{Hyperbolic tangent activation}
Consider a neuron with input
$\bm{x}$, weights $\bm{w}$, a hyperbolic tangent activation function, and a
stochastic output $y\in\{-1,1\}$, given by
\begin{equation}
p(y=1) = \frac{1+ \tanh( \bm{w}^\intercal \tilde{\bm{x}} )}{2},
\quad \tanh(t)=\frac{\exp(t)-\exp(-t)}{\exp(t)+\exp(-t)}.
\end{equation}
For convenience, throughout this paper
$\tilde{\bm{x}}=(\bm{x}^\intercal,1)^\intercal$ denotes the augmented vector of
$\bm{x}$ (homogeneous coordinates) so that $\bm{w}^\intercal \tilde{\bm{x}}$ contains a bias term, and a
general linear transformation can be written simply as $\bm{A}\tilde{\bm{x}}$.
By \cref{def:rfim} and some simple analysis~\footnote{See
the appendix for detailed derivations.}, we get
\begin{equation}\label{eq:fimtanh}
g^{y}( \bm{w}\,\vert\,\bm{x} )
=
\nu_{\tanh}( \bm{w}, \bm{x} )\tilde{\bm{x}}\tilde{\bm{x}}^\intercal,
\quad
\nu_{\tanh}( \bm{w}, \bm{x} )
= 1 - \tanh^2( \bm{w}^\intercal \tilde{\bm{x}} ).
\end{equation}

\subsubsection{Sigmoid activation}
Similarly, the RFIM of a neuron with input $\bm{x}$, weights $\bm{w}$, a sigmoid
activation function, and a stochastic binary output $y\in\{0,1\}$, where
\begin{equation}
p(y=1)=\sigm(\bm{w}^\intercal \tilde{\bm{x}}),
\quad
\sigm(t)=\frac{1}{1+\exp(-t)}, 
\end{equation}
is given by
\begin{align}\label{eq:fimsigm}
g^y(\bm{w}\,\vert\,\bm{x})
&=
\nu_{\sigm}(\bm{w},\bm{x}) \tilde{\bm{x}}\tilde{\bm{x}}^\intercal,\nonumber\\
\nu_{\sigm}(\bm{w},\bm{x})
&=
\sigm\left( \bm{w}^\intercal \tilde{\bm{x}} \right)
\big[ 1 - \sigm\left( \bm{w}^\intercal \tilde{\bm{x}} \right) \big].
\end{align}
A neuron with sigmoid activation but continuous output was discussed earlier
\citep{amari96}.

\subsubsection{Parametric Rectified Linear Unit}

Another commonly used activation function is Parametric Rectified Linear Unit
(PReLU)~\citep{he15}, which includes Rectified Linear Unit (ReLU)~\citep{nair10}
as a special case. To compute the RFIM, we formulate PReLU into a conditional
distribution given by
\begin{equation}\label{eq:relu}
p( y\,\vert\,\bm{w}, \bm{x} )
=
G( y\,\vert\, \relu(\bm{w}^\intercal\tilde{\bm{x}}), \sigma^2 ),
\quad
\relu(t) = 
\left\{
\begin{array}{ll}
t        & \text{if }t \ge 0\\
\iota t & \text{if }t < 0.
\end{array}
\right.
\quad
(0\le\iota<1)
\end{equation}

When $\iota=0$, \cref{eq:relu} becomes ReLU. 
By \cref{def:rfim}, the corresponding RFIM is
\begin{equation}\label{eq:fimrelunonsmooth}
g^y(\bm{w}\,\vert\,\bm{x}) = \left\{
\begin{array}{ll}
\frac{1}{\sigma^2}\tilde{\bm{x}}\tilde{\bm{x}}^\intercal       & \text{if }\bm{w}^\intercal\tilde{\bm{x}}>0\\
\text{undefined}               & \text{if }\bm{w}^\intercal\tilde{\bm{x}}=0\\
\frac{\iota^2}{\sigma^2}\tilde{\bm{x}}\tilde{\bm{x}}^\intercal & \text{if }\bm{w}^\intercal\tilde{\bm{x}}<0
\end{array}
\right.
\end{equation}
This RFIM is discontinuous at $\bm{w}^\intercal\tilde{\bm{x}}=0$. To
obtain a smoother RFIM, a trick is to replace $\relu(\bm{w}^\intercal\tilde{\bm{x}})$ on
the left-hand-side of \cref{eq:relu} with
$\relu_\omega(\bm{w}^\intercal\tilde{\bm{x}})$, where
\begin{equation}
\relu_\omega(t)=\omega \ln\left(
\exp\left(\frac{\iota{t}}{\omega}\right)
+
\exp\left(\frac{t}{\omega}\right)
\right),
\end{equation}
and $\omega>0$ is a hyper-parameter so that $\lim_{\omega\to0^+}
\relu_\omega=\mathtt{relu}$.  Then, PReLU's RFIM is given by
\begin{align}\label{eq:fimrelu}
g^y(\bm{w}\,\vert\,\bm{x})
&=
\nu_{\relu}(\bm{w},\bm{x})
\tilde{\bm{x}}\tilde{\bm{x}}^\intercal,\nonumber\\
\nu_{\relu}(\bm{w},\bm{x})
&=
\frac{1}{\sigma^2}
\left[
\iota + (1-\iota) \sigm\left(
\frac{1-\iota}{\omega}
\bm{w}^\intercal\tilde{\bm{x}}
\right)
\right]^2,
\end{align}
which is simply a smoothed version of \cref{eq:fimrelunonsmooth}.
If we set empirically $\sigma=1$, $\iota=0$, then $\nu_{\relu}(\bm{w},\bm{x})=
\sigm^2\left( \frac{1}{\omega} \bm{w}^\intercal\tilde{\bm{x}} \right)$ is close
to 1 when $\bm{w}^\intercal\tilde{\bm{x}}>0$, and is close to 0 otherwise.

\subsubsection{Exponential Linear Unit}

A stochastic exponential linear unit (ELU)~\citep{clevert15} with $\alpha>0$ is
\begin{equation}
p(y\,\vert\,\bm{w},\,\bm{x})
=
G(y\,\vert\,\elu(\bm{w}^\intercal\tilde{\bm{x}}),\,\sigma^2),
\quad
\elu(t)
=\left\{
\begin{array}{ll}
t & \text{if }t\ge0\\
\alpha \left( \exp(t) - 1 \right) & \text{if }t<0.\\
\end{array}
\right.
\end{equation}
Its RFIM is given by
\begin{align}
g^y(\bm{w}\,\vert\,\bm{x})
&= \nu_{\elu}(\bm{w},\bm{x})
\tilde{\bm{x}}\tilde{\bm{x}}^\intercal,\nonumber\\
\nu_{\elu}(\bm{w},\bm{x})
&=
\left\{
\begin{array}{ll}
\frac{1}{\sigma^2} & \text{if }\bm{w}^\intercal\tilde{\bm{x}}\ge0\\
\frac{\alpha^2}{\sigma^2}\exp\left( 2\bm{w}^\intercal\tilde{\bm{x}} \right)
& \text{if }\bm{w}^\intercal\tilde{\bm{x}}<0.\\
\end{array}
\right.
\end{align}
The coefficient function $\nu_{\elu}(\bm{w},\bm{x})$ is continuous 
but non-differentiable at $\bm{w}^\intercal\tilde{\bm{x}}=0$.

\subsubsection{A generic expression of one-neuron RFIMs}

Denote $f\in\{ \tanh, \sigm, \relu, \elu \}$ to be an element-wise nonlinear
activation function.  By \cref{eq:fimtanh,eq:fimsigm,eq:fimrelu}, the RFIMs of
single neurons have a common form
\begin{equation}
g^y(\bm{w}\,\vert\,\bm{x})
= \nu_f(\bm{w},\bm{x})\tilde{\bm{x}}\tilde{\bm{x}}^\intercal,
\end{equation}
where $\nu_f(\bm{w},\bm{x})$ is a positive coefficient with large values in the
linear region, or the effective learning zone of the neuron.

\subsection{RFIM of One Layer}

A \emph{linear} layer with input $\bm{x}$, connection weights
$\bm{W}=[\bm{w}_1,\cdots,\bm{w}_{D_y}]$, and stochastic output
$\bm{y}$ can be represented by $p(
\bm{y}\,\vert\,\bm{W}, \bm{x} ) = G( \bm{y}\,\vert\,\bm{W}^\intercal
\tilde{\bm{x}}, \sigma^2 \bm{I} )$, where $\bm{I}$ is the identity matrix, and
$\sigma$ is the scale of the observation noise. We vectorize $\bm{W}$ by
stacking its columns $\{\bm{w}_i\}$, then $g^{\bm{y}}(\bm{W}\,\vert\,\bm{x})$ is
a tensor of size $(D_x+1)D_y\times(D_x+1)D_y$, where $D$ denotes the dimension
of the corresponding variable.  Fortunately, due to conditional independence of
$\bm{y}$'s dimensions given $\bm{W}$ and $\bm{x}$, the RFIM has a simple block
diagonal form, given by
\begin{equation}\label{eq:linear}
g^{\bm{y}}(\bm{W}\,\vert\,\bm{x})
=
\frac{1}{\sigma^2}
\diag\left[
\tilde{\bm{x}}\tilde{\bm{x}}^\intercal,
\cdots,
\tilde{\bm{x}}\tilde{\bm{x}}^\intercal
\right],
\end{equation}
where $\mathtt{diag}(\cdot)$ means the (block) diagonal matrix constructed by
the given matrix entries.

A \emph{nonlinear} layer increments a linear layer by adding an element-wise
activation function applied on $\bm{W}^\intercal\tilde{\bm{w}}$, and randomized
wrt the choice of the activation (Bernoulli for $\tanh$ and $\sigm$; Gaussian
for $\relu$). By \cref{def:rfim}, its RFIM is given by
\begin{equation}\label{eq:1layer}
g^{\bm{y}} \left( \bm{W}\,\vert\,\bm{x} \right)
=
\mathtt{diag}\left[\,
\nu_f(\bm{w}_1,\bm{x})\tilde{\bm{x}}\tilde{\bm{x}}^\intercal,%
\,\cdots,\,%
\nu_f(\bm{w}_m,\bm{x})\tilde{\bm{x}}\tilde{\bm{x}}^\intercal
\,\right],
\end{equation}
where $\nu_f(\bm{w}_i,\bm{x})$ depends on the activation function $f$
as discussed in \cref{subsec:1neuron}.

A softmax layer, which often appears as the last layer of a MLP, is given by
$y\in\{1,\dots,m\}$, where 
\begin{equation}
p(y) =
\eta_y =
\frac{\exp(\bm{w}_y\tilde{\bm{x}})}
{ \sum_{i=1}^m \exp(\bm{w}_i\tilde{\bm{x}}) }.
\end{equation}
Its RFIM is not block diagonal any more but given by
\begin{equation}
g^{\bm{y}}(\bm{W})
=
\begin{bmatrix}
(\eta_1 - \eta_1^2) \tilde{\bm{x}} \tilde{\bm{x}}^\intercal & 
-\eta_1\eta_2  \tilde{\bm{x}} \tilde{\bm{x}}^\intercal & 
\cdots & - \eta_1\eta_m \tilde{\bm{x}} \tilde{\bm{x}}^\intercal \\
-\eta_2\eta_1  \tilde{\bm{x}} \tilde{\bm{x}}^\intercal & 
(\eta_2 - \eta_2^2) \tilde{\bm{x}} \tilde{\bm{x}}^\intercal 
&\cdots&
- \eta_2\eta_m \tilde{\bm{x}} \tilde{\bm{x}}^\intercal \\
\vdots & \vdots &\ddots & \vdots \\
-\eta_m\eta_1 \tilde{\bm{x}} \tilde{\bm{x}}^\intercal &
-\eta_m\eta_2 \tilde{\bm{x}} \tilde{\bm{x}}^\intercal &
\cdots &
(\eta_m - \eta_m^2) \tilde{\bm{x}} \tilde{\bm{x}}^\intercal \\
\end{bmatrix}.
\end{equation}
Notice that its $i$'th diagonal block $(\eta_i - \eta_i^2) \tilde{\bm{x}}
\tilde{\bm{x}}^\intercal$ resembles the RFIM of a single $\sigm$ neuron.

\subsection{RFIM of Two Layers}

\begin{figure}[t]
\centering
\includegraphics[width=.3\textwidth]{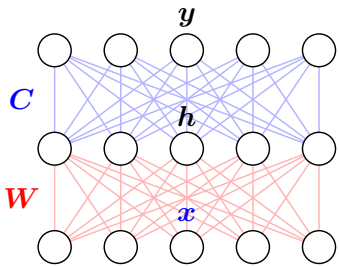}
\caption{Two layers embedded in a MLP. The red lines and symbols show the
interested parameters whose RFIM is to be computed. The blue lines and symbols
show the reference variables that is considered given.}\label{fig:twolayer}
\end{figure}

By \cref{eq:1layer}, the one-layer RFIM is a product metric~\citep{jost11} and
does not consider the inter-neuron correlations. We must look at a larger
subsystem to obtain such correlations. Consider a two-layer model with
stochastic output $\bm{y}$ around the mean vector $f\left(\bm{C}^\intercal
\bm{h}\right)$, where $\bm{h}=f\left(\bm{W}^\intercal \bm{x}\right)$, as 
shown in \cref{fig:twolayer}.
For simplicity, we ignore inter-layer correlations between the first layer and
the second layer and focus on the inter-neuron correlations within the first layer.
To do this, both $\bm{x}$ and $\bm{C}$ are considered as references to compute
the RFIM of $\bm{W}$. By \cref{def:rfim},
$g^{\bm{y}}(\bm{W}\,\vert\,\bm{x},\bm{C})= \left[ \bm{G}_{ij}
\right]_{D_h\times{}D_h}$ and each block
\begin{equation}\label{eq:twolayer}
\bm{G}_{ij}=
\sum_{l=1}^{D_y}
c_{il} c_{jl}
\nu_f( \bm{c}_l, \bm{h} )
\nu_f( \bm{w}_i, \bm{x} )
\nu_f( \bm{w}_j, \bm{x} )
\tilde{\bm{x}} \tilde{\bm{x}}^\intercal,
\quad
\forall 1 \le i \le D_h,
\quad
\forall 1 \le j \le D_h.
\end{equation}
Consider the computational difficulty of \cref{eq:twolayer} that is only listed here as
an analytical contribution with possible empirical extensions.  

\section{Advantages of RFIM}\label{sec:norm}

This section discusses theoretical advantages of RFIM. Consider wlog a MLP
with one Bernoulli output $y$, whose mean $\mu$ is a deterministic function
depending on the input $\bm{x}$ and the network parameters $\bm\Theta$. By
\cref{sec:review}, the FIM of the MLP can be computed as
\begin{equation}
g(\bm\Theta)
= 
\overline{
\mu_i \frac{\partial\ln\mu_i}{\partial\bm\Theta}
\frac{\partial\ln\mu_i}{\partial\bm\Theta^\intercal}
+ (1-\mu_i) \frac{\partial\ln(1-\mu_i)}{\partial\bm\Theta} 
\frac{\partial\ln(1-\mu_i)}{\partial\bm\Theta^\intercal} }
=
\overline{
\frac{1}{\mu_i(1-\mu_i)}
\frac{\partial\mu_i}{\partial\bm\Theta}
\frac{\partial\mu_i}{\partial\bm\Theta^\intercal} }.
\end{equation}
Therefore $\mathtt{rank}(g(\bm\Theta))\le{n}$, as each sample contributes
maximally 1 to the rank of $g(\bm\Theta)$. A small diagonal block of
$g(\bm\Theta)$, representing one layer, is likely to have a rank much lower than
the sample size. If the number of parameters is greater than the sample size,
which can be achieved especially with deep learning~\citep{szegedy15}, then
$g(\bm\Theta)$ is guaranteed to be singular. All methods trying to approximate
FIM suffers from this problem and should use proper regularizations.
Comparatively, RFIM decomposes the network in a layer-wise manner. In each layer
$\bm{h}=f(\bm{W}^\intercal\bm{x})$, by \cref{eq:1layer},
$\mathtt{rank}(g^{\bm{h}}(\bm{W}))\le{n}D_h$, which is an achievable upper bound.
Therefore,
RFIM is expected to have a much higher rank than FIM. Higher rank means less
singularity, and more information is captured, and easier to reparameterize to
achieve good optimization properties. Essentially, RFIM integrates the
\emph{internal stochasticity}~\citep{bengio13} of the neural system by
considering the output of each layer as random variables. In theory, the
computation of FIM should also consider stochastic neurons.  However it requires
to marginalize the joint distribution of $\bm{h}_1$, $\bm{h}_2$, $\cdots$,
$\bm{y}$. This makes the already infeasible computation even more difficult. 

RFIM is accurate, for that the geometry of $\bm\theta$ is defined wrt to its
direct response $\bm{h}$ in the system, or adjacent nodes in a graphical model.
By the example in \cref{sec:review} and \cref{eq:rfim}, the RFIM
$g^y(\bm\theta_{L})$ is exactly the corresponding block in the FIM
$I(\bm\Theta)$, because they both investigate how $\bm\theta_L$ affects the last
layer $\bm{h}_{L-1}\to{y}$. They start to diverge from the second last layer.
To compute the geometry of $\bm\theta_{L-1}$, RFIM looks at how
$\bm\theta_{L-1}$ affects the local mapping $\bm{h}_{L-1}\to\bm{h}_L$.
Intuitively, this local relationship can be more reliably measured regardless of
the rest of the system. In contrast, FIM examines how $\bm\theta_{L-1}$ affects
the global mapping $\bm{h}_{L-1}\to{y}$. This task is much more difficult,
because it must consider the correlation between different layers. This is hard
to perform without approximation techniques. As a commonly used approximation,
the block diagonalized version of FIM will ignore such correlations and loose
accuracy.

The measurement of RFIM makes it possible to maintain global stability of the
learning system by balancing different subsystems. Consider two connected
subsystems with internal parameters $\bm\theta_1$ and $\bm\theta_2$ and
corresponding responses $\bm{h}_1$ and $\bm{h}_2$.  A learning step is given by
$\bm\theta_1\leftarrow\bm\theta_1+\delta\bm\theta_1$ and
$\bm\theta_2\leftarrow\bm\theta_2+\delta\bm\theta_2$, with
$\Vert\delta\bm\theta_1\Vert\le\gamma$ and
$\Vert\delta\bm\theta_2\Vert\le\gamma$ constrained by a pre-chosen maximum radius
$\gamma$. To balance the system, we constrain $g^{\bm{h}_1}(\bm\theta_1)$ and
$g^{\bm{h}_2}(\bm\theta_2)$ to have similar scales, e.g. by normalizing their
largest eigenvalue to 1. This can be done through reparametrization tricks.
Then the intrinsic changes, or variations, of the responses $\bm{h}_1$ and
$\bm{h}_2$ also have similar scales and is upper bounded. Note that the
responses $\bm{h}_1$ and $\bm{h}_2$ are the interfaces of subsystems. For
example, in the MLP shown in \cref{fig:geometry}, the response $\bm{h}_1$ in
subsystem 1 becomes the reference in subsystem 2. By bounding the variation of
$\bm{h}_1$, the Riemannian metric $g^{\bm{h_2}}(\bm\theta_2\,\vert\,\bm{h}_1)$
will present less variations during learning.

\section{Relative Natural Gradient Descent (RNGD)}\label{sec:rng}

The traditional non-parametric way of applying natural gradient requires to
re-calculate the FIM and solving a large linear system in each learning step.
Besides the huge computational cost, it meets some difficulties. For example,
in an online learning scenario, a mini batch of samples cannot faithfully reflect
the ``true'' geometry, which has to integrate the risk of sample variations.
That is, the FIM of a mini batch is likely to be singular or with bad
conditions.

A recent series of efforts~\citep{montavon12,raiko12,desjardins15} are gearing towards a
parametric approach of applying natural gradient, which memorizes and
\emph{learns a geometry}. For example, natural neural
networks~\citep{desjardins15} augment each layer with a redundant linear layer,
and let these linear layers to parametrize the geometry of the neural manifold.

By dividing the learning system into subsystems, RFIM makes this parametric
approach much more implementable. The memory complexity of storing the
Riemannian metric has been reduced from $O(\#\bm\Theta^2)$ to
$O(\sum_{i}\#\bm\theta_i^2)$, where each $\bm\theta_i$ corresponds to a
subsystem, and $\#\bm\Theta$ means the dimensionality of $\bm\Theta$.
The computational complexity has been reduced from
$O(\#\bm\Theta^\varrho)$ ($\varrho\approx2.373$, \cite{williams2012}) to
$O(\sum_{i}\#\bm\theta_i^\varrho)$. Approximation techniques of FIM to
improve these complexities can be applied to RFIM. Optimization based on RFIM
will be called Relative Natural Gradient Descent (RNGD).  In the following of
this section, we present two examples of RNGD. The objective is to demonstrate
its advantages and mechanisms.

\subsection{RNGD with a Single $\sigm$ Neuron}

The first experiment is a single neuron model to implement logistic
regression~\citep{minka03}. The focus of the experiment is to demonstrate the
mechanisms of RNGD and to show on a small scale model the improvement made by RNGD
over feature whitening techniques, so that we can expect the same improvement on
a larger model.

To classify a sample set $\{(\bm{x}_i,y_i)\}$ with features $\bm{x}$ and class
labels $y\in\{0,1\}$, a statistical model is assumed as
\begin{align}\label{eq:rngd1}
p(y=1) &
=\sigm( \bm{\theta}^\intercal \bm{z} ),\nonumber\\
\bm{z} &= ((\bm{x}-\bm{a})^\intercal \bm{A}^\intercal, 1 )^\intercal.
\end{align}
In \cref{eq:rngd1}, $\bm\theta$ is the canonical parameters or the link weights
of the neuron. $\bm{A}$ and $\bm{a}$ are for feature
whitening~\citep{montavon12,desjardins15}. They are precomputed
and fixed during learning, so that the transformed samples
$\{\bm{A}(\bm{x}_i-\bm{a})\}$ have zero mean and unit covariance except singular
dimensions. The learning cost function is the average cross entropy
\begin{equation}
L(\bm\theta) =
\overline{
- y_i \ln \sigm( \bm{\theta}^\intercal \bm{z}_i )
- (1-y_i) \ln \left[ 1-\sigm( \bm{\theta}^\intercal \bm{z}_i ) \right]},
\end{equation}
whose gradient is simply 
$
\bigtriangledown{L} =
\overline{
\left(
\sigm( \bm{\theta}^\intercal \bm{z}_i ) - y_i
\right)
\bm{z}_i}
$. To apply RNGD, by \cref{eq:fimsigm}, we have
$g^y(\bm\theta) =
\overline{ 
\nu_\sigm\left( \bm\theta, \bm{z}_i \right)
\bm{z}_i \bm{z}_i^\intercal }$.
In each learning step, we update $\bm\theta$ based on
\begin{equation}\label{eq:logisticupdate}
\bm\theta^{\mathrm{new}} \leftarrow
\bm\theta^{\mathrm{old}}
-
\gamma ( g^y(\bm\theta) + \epsilon \bm{I} )^{-1} \bigtriangledown{L},
\end{equation}
where $\gamma$ is a learning rate, and $\varepsilon>0$ is a hyper-parameter to
guarantee that $\left(g^y(\bm\theta) + \epsilon \bm{I}\right)$ is invertible.  We
choose empirically $\epsilon$ to be $\varepsilon \mathrm{tr}(g^y(\bm\theta))/D$, where
$\varepsilon=10^{-2}$.

Based on a gradient descent optimizer with constant learning rate and momentum,
we compare four different methods: 
\begin{enumerate}
\item \texttt{GD} fixes $\bm{A}=\bm{I}$,
$\bm{a}=\bm0$, and applies gradient descent; 
\item \texttt{WhiteGD} performs feature
whitening by pre-computing $\bm{A}$ and $\bm{a}$ as well as gradient descent;
\item\texttt{NGD} fixes $\bm{A}=\bm{I}$, $\bm{a}=\bm0$, and updates $\bm\theta$ based
on \cref{eq:logisticupdate}; 
\item\texttt{WhiteNGD} performs both feature whitening
and the updating scheme in \cref{eq:logisticupdate}.
\end{enumerate}

\begin{figure}[t!]
\centering
\begin{subfigure}[b]{.45\textwidth}
\includegraphics[width=\textwidth]{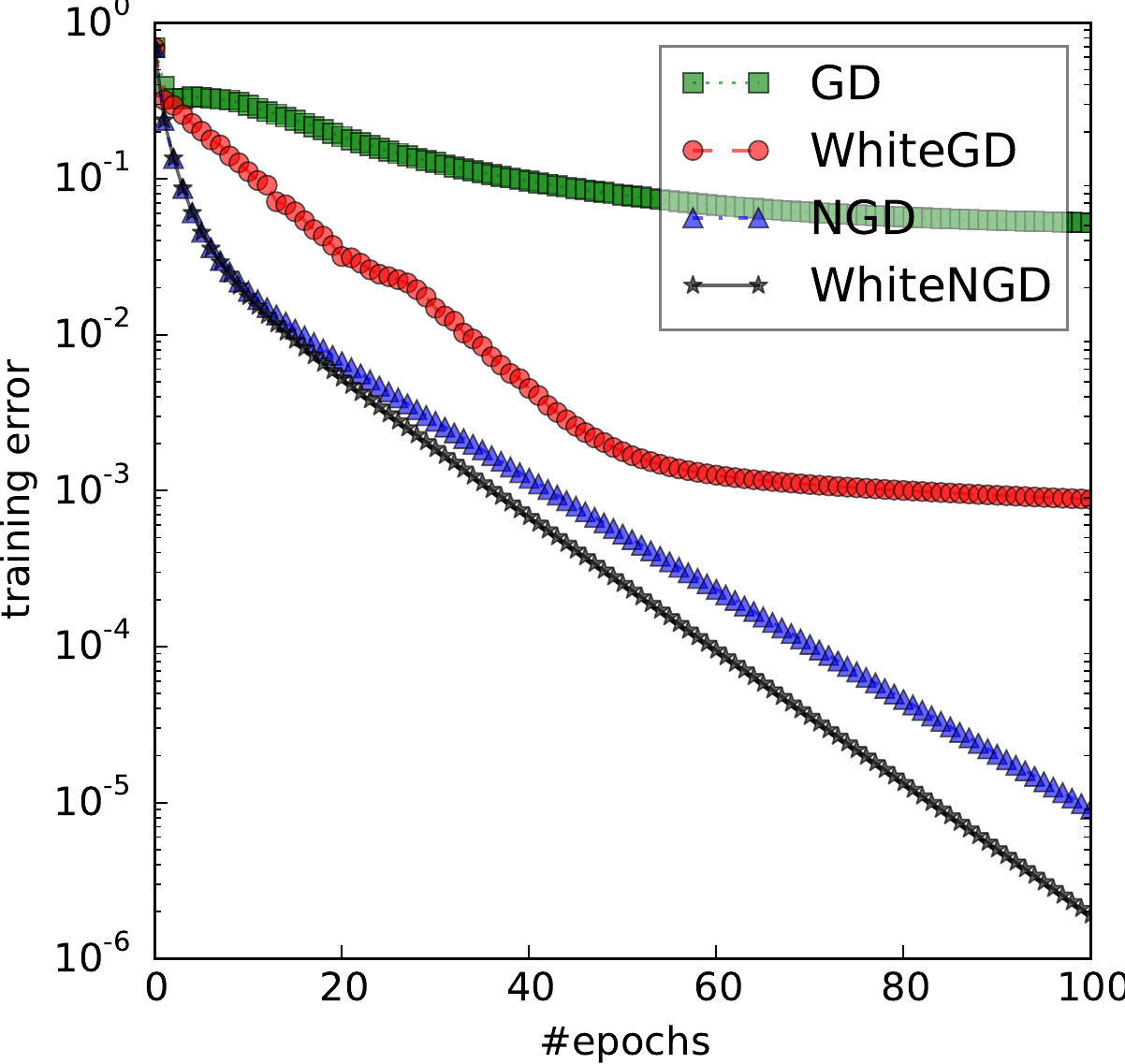}
\caption{``3'' vs ``5''; 50\% for training}
\end{subfigure}
\begin{subfigure}[b]{.45\textwidth}
\includegraphics[width=\textwidth]{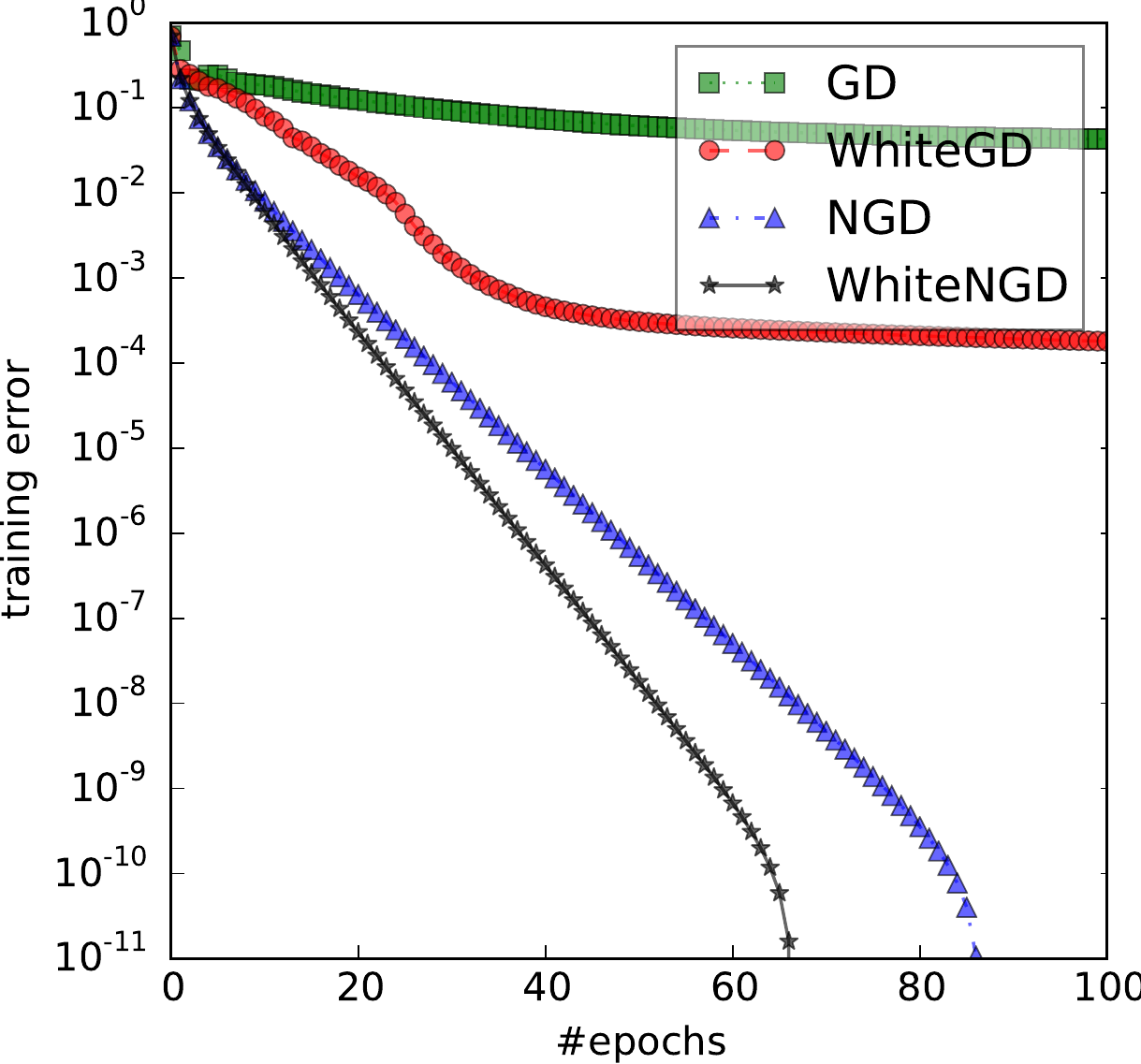}
\caption{``4'' vs ``9''; 50\% for training}
\end{subfigure}
\begin{subfigure}[b]{.45\textwidth}
\includegraphics[width=\textwidth]{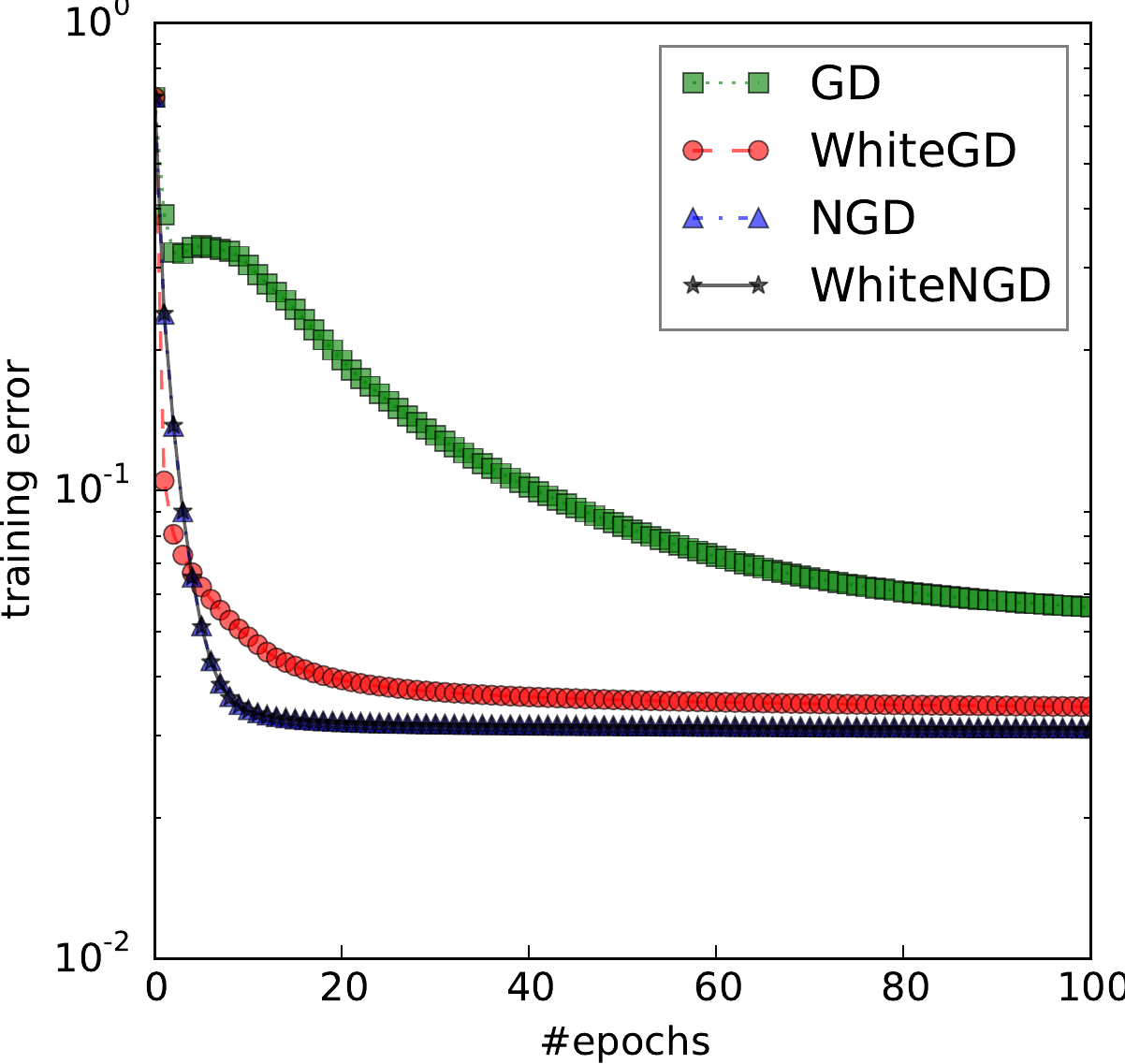}
\caption{``3'' vs ``5''; 60\% for training}\label{fig:whitefast}
\end{subfigure}
\begin{subfigure}[b]{.45\textwidth}
\includegraphics[width=\textwidth]{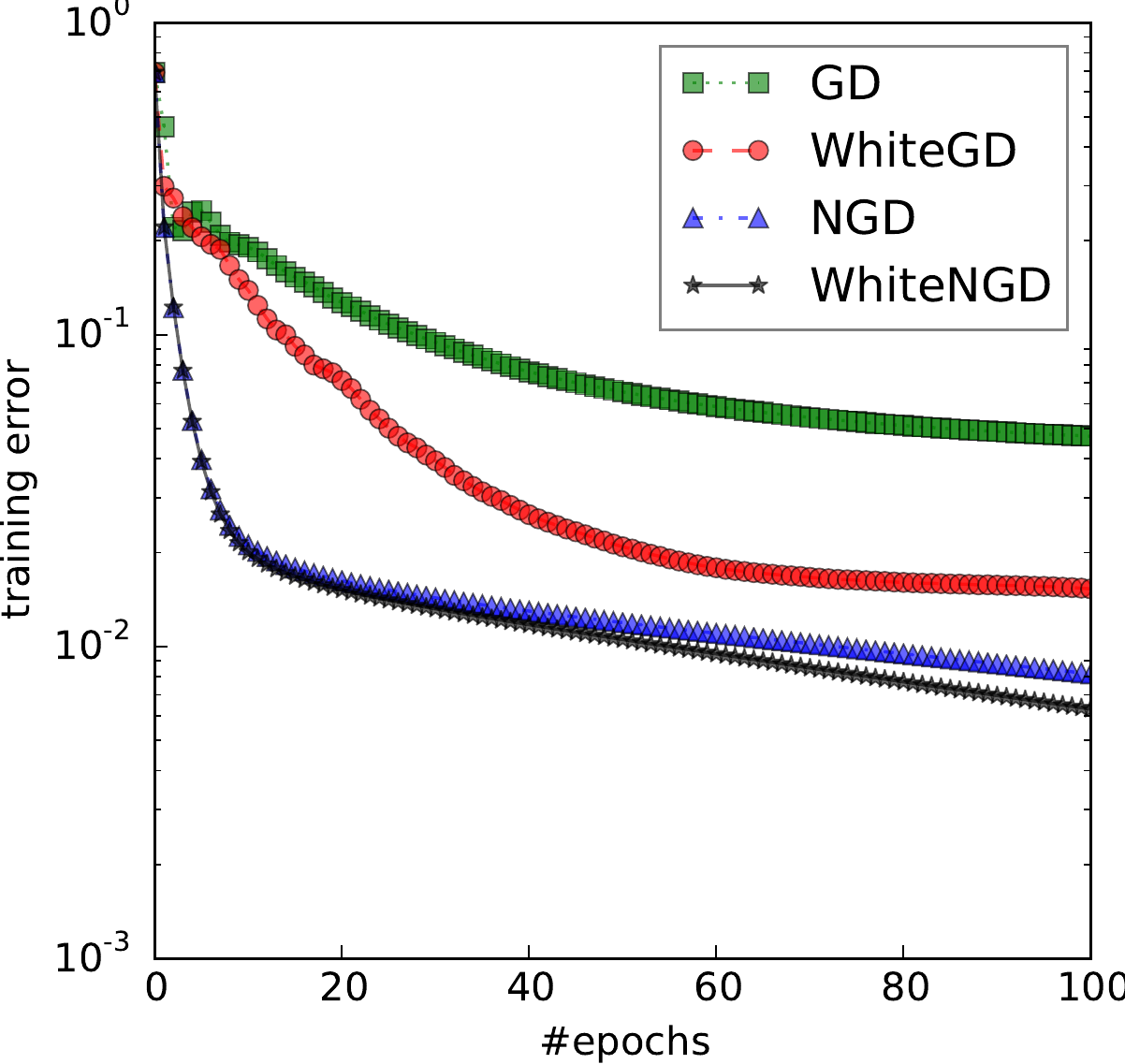}
\caption{``4'' vs ``9''; 60\% for training}
\end{subfigure}
\caption{Learning curves of single $\sigm$ neuron classifiers (better viewed in
color). Each curve is an average over ten runs. For each method, the best curve
over a configuration grid of learning rates and momentums is presented.}\label{fig:toy}
\end{figure}

Based on the MNIST dataset~\citep{mnist}, \cref{fig:toy} shows their learning
curves on binary classification problems ``3'' vs ``5'' and ``4'' vs ``9'' on
two different training sizes. For each method, the best curve which achieved
the minimal training error after 100 epochs is shown. The configuration grid is
given as follows. The candidate learning rate is in the range $\{ 10^{-2},
10^{-1}, 1, 10, 100 \}$.  The candidate momentum is in the range $\{0,0.8\}$.

In different cases, \texttt{WhiteNGD} and \texttt{NGD} can consistently reach
deeper in the error surface within a reasonable number of epochs as compared
to \texttt{WhiteGD} and \texttt{GD}. Among all methods, \texttt{WhiteNGD}
performs best, which demonstrates the dependency of natural gradient and RFIM on
the coordinate system. In a whitened coordinate system, RFIM is expected to have
better conditions in average, and therefore leading to better optimization.
Intuitively, in \cref{eq:fimsigm}, the term $\nu_\sigm\left( \bm\theta, \bm{x}
\right)$ serves as a ``selector'', highlighting a subset of $\{\bm{x}_i\}$ in
the linear region of the perceptron.  The updating rule in
\cref{eq:logisticupdate} will zoom in and de-correlate the sample variance in
this region, and let the classifier focus on the discriminative samples. 

Gradient descent depends much more on the choice of the coordinate system. There
is a significant improvement from \texttt{GD} to \texttt{WhiteGD}. A whitened
coordinate system allows larger learning rates. In the first several epochs,
\texttt{WhiteGD} can even learn faster than \texttt{NGD} and \texttt{WhiteNGD},
as shown in \cref{fig:whitefast}.


\subsection{RNGD with a $\relu$ MLP}

The good performance of batch normalization (BN)~\citep{ioffe15} can be
explained using RFIM. Basically, BN uses an \emph{inter-sample} normalization
layer to transform the input of each layer, denoted as $\bm{x}$, to be zero mean
and unit variance and thus reduces ``internal covariate shift''. In a typical
case, above this normalization layer is a linear layer given by
$\bm{y}=\bm{W}^\intercal\bm{x}$. By \cref{eq:linear}, if $\bm{x}$ is normalized,
then the diagonal entries of $g^{\bm{y}}(\bm{W})$ becomes uniform. Therefore the
geometry of the parameter manifold is conditioned. In this case, BN helps to
condition the RFIM of a linear layer.

This subsection looks at a larger subsystem consisting of a linear layer plus a
non-linear activation layer above it, given by
$\bm{y}=f(\bm{W}^\intercal\bm{x})$. By \cref{eq:1layer}, its RFIM
is 
$\mathtt{diag}\left[\,%
\nu_f(\bm{w}_1,\bm{x})
\tilde{\bm{x}}\tilde{\bm{x}}^\intercal,
\,\cdots,\,
\nu_f(\bm{w}_m,\bm{x})
\tilde{\bm{x}}\tilde{\bm{x}}^\intercal\,\right]$. To perform RNGD,
one need to update this layer by
\begin{align}
\bm{w}_1^{\mathrm{new}}
& \leftarrow \bm{w}_1^{\mathrm{old}}
-
\overline{
\nu_f(\bm{w}_1,\bm{x}_i)
\tilde{\bm{x}}_i\tilde{\bm{x}}_i^\intercal
+ \epsilon \bm{I}
}^{-1}
\frac{\partial{E}}{\partial\bm{w}_1},\nonumber\\
\cdots\nonumber\\
\bm{w}_m^{\mathrm{new}} 
&\leftarrow \bm{w}_m^{\mathrm{old}}
-
\overline{
\nu_f(\bm{w}_m,\bm{x}_i)
\tilde{\bm{x}}_i\tilde{\bm{x}}_i^\intercal
+ \epsilon \bm{I} }^{-1}
\frac{\partial{E}}{\partial\bm{w}_m},
\end{align}
where $E$ is the cost function, and $\epsilon>0$ is a hyper parameter to avoid
singularity. However, this update requires solving many linear subsystems and is
too expensive to compute. Moreover, we only have a mini batch which contains not
enough information to compute the RFIM.  To tackle these difficulties, we
maintain an \emph{exponentially moving average} of $\bm{G}_l$, the RFIM of the
$l$'th neuron in this layer. Initially, $\bm{G}_l$ is initialized to identity.
At each iteration, it is updated by
\begin{equation}
\bm{G}_l^{\mathrm{new}}
\leftarrow 
\lambda \bm{G}_l^{\mathrm{old}}
+ (1-\lambda)
\overline{
\nu_f(\bm{w}_l,\bm{x}_i)
\tilde{\bm{x}}_i\tilde{\bm{x}}_i^\intercal
+ \epsilon \bm{I}},
\end{equation}
where the average is taken over all samples in this mini batch, and $\lambda$ is
a decaying rate. Every $T$ iterations, we recompute $\bm{G}_l^{-1}$ based on the
most current $\bm{G}_l$, and store the resulting $\bm{G}_l^{-1}$. In the next
$T$ iterations, this $\bm{G}_l^{-1}$ will remain constant and be used as an
approximation of inverse RFIM. Then, the updating rule of the layer is given by 
\begin{equation}
\bm{w}_1^{\mathrm{new}} \leftarrow \bm{w}_1^{\mathrm{old}}
- \bm{G}_1^{-1} \frac{\partial{E}}{\partial\bm{w}_1}
\quad\cdots\quad
\bm{w}_m^{\mathrm{new}} \leftarrow \bm{w}_m^{\mathrm{old}}
- \bm{G}_m^{-1} \frac{\partial{E}}{\partial\bm{w}_m}.
\end{equation}
For the input layer which scales with the number of input features, and the
final soft-max layer, we apply instead the RFIM of the corresponding linear
layer for the consideration of the computational efficiency.

\begin{figure}[t!]
\centering
\begin{minipage}[c]{0.3\textwidth}%
\includegraphics[width=\textwidth]{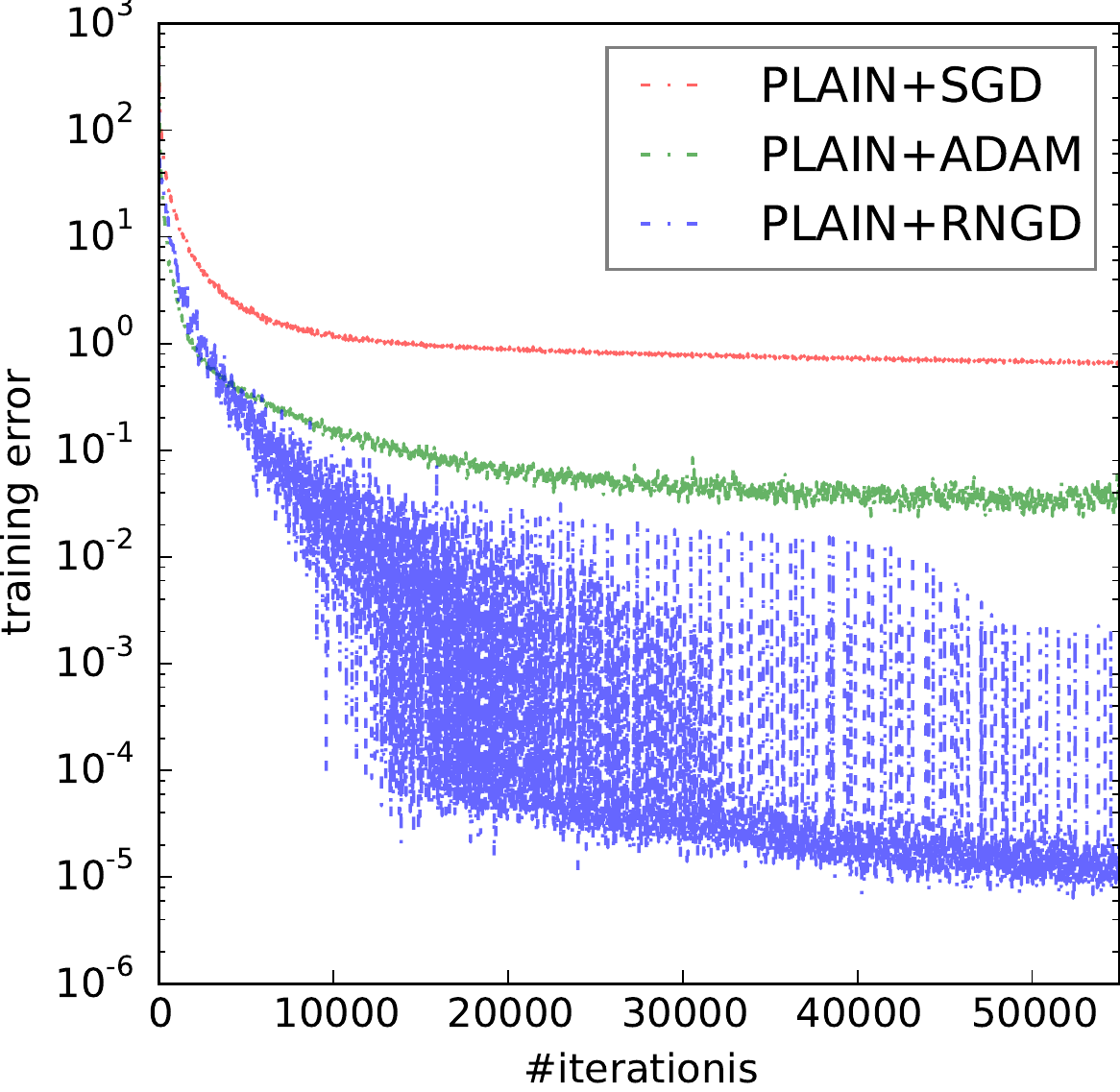}\\
\includegraphics[width=\textwidth]{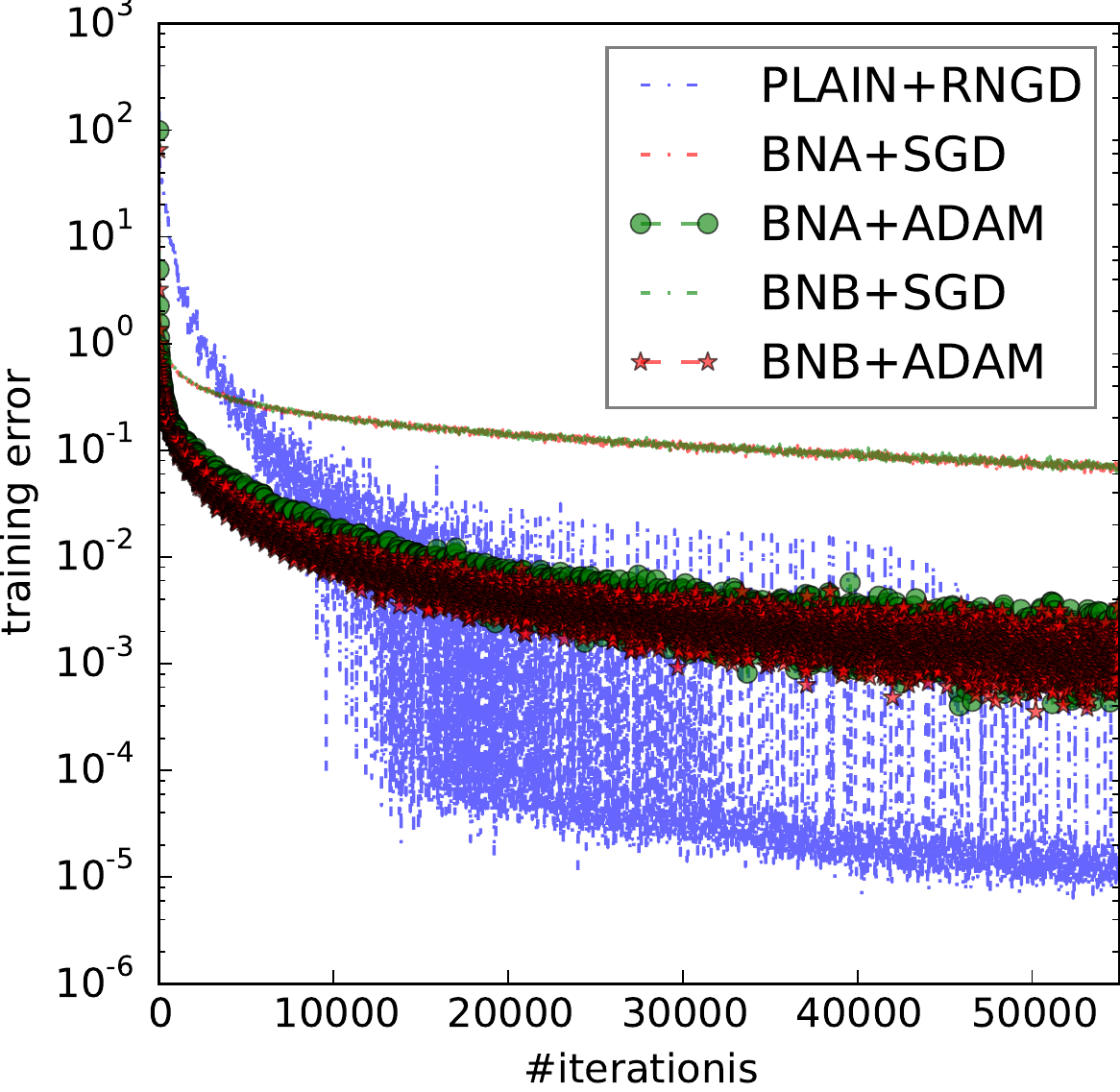}
\end{minipage}
\begin{minipage}[c]{0.65\textwidth}%
\centering
\small
\begin{tabular}{rcccc}
\hline
& PLAIN+SGD & PLAIN+ADAM & PLAIN+RNGD \\
\hline
$\tau=0.1$ & $0.67\pm0.3$ & $0.035\pm0.04$ & $5\times10^{-5}\pm0.002$  \\
$\tau=0.5$ & $0.72\pm0.3$ & $0.039\pm0.04$ & $2.8\times10^{-4}\pm0.01$ \\
\hline
$\gamma$ & $10^{-3}$ & $10^{-2}$ & $10^{-2}$ \\
\hline
\end{tabular}
\vspace{20pt}

\begin{tabular}{rcc}
\hline
& BNA+SGD & BNA+ADAM \\
\hline
$\tau=0.1$ & $0.072\pm0.04$ & $0.0015\pm0.005$ \\
$\tau=0.5$ & $0.089\pm0.05$ & $0.0021\pm0.006$ \\
\hline
$\gamma$ & $0.1$ & $10^{-2}$ \\
\hline
\end{tabular}
\vspace{20pt}

\begin{tabular}{rcc}
\hline
& BNB+SGD & BNB+ADAM \\
\hline
$\tau=0.1$ & $0.073\pm0.04$ & $0.0013\pm0.004$ \\
$\tau=0.5$ & $0.090\pm0.05$ & $0.0018\pm0.005$ \\
\hline
$\gamma$   & $0.1$ & $10^{-2}$ \\
\hline
\end{tabular}
\end{minipage}
\caption{Learning curves on MNIST with $\relu$ MLPs (better viewed in color).
For each method, the best learning curve over several configurations of learning
rates is selected. The curves are smoothed by averaging over every 10 iterations
for a clear visualization.} \label{fig:toy2}
\end{figure}

We implemented the proposed method using TensorFlow~\citep{tensorflow} and
applied it to classify MNIST digits. The network has shape 784-64-64-64-10,
with $\relu$ activation units, a final soft-max layer, and uses average
cross-entropy as the cost function. 

\Cref{fig:toy2} shows the learning curves of different methods. \texttt{SGD} is
stochastic gradient descent. \texttt{ADAM} is the Adam optimizer~\citep{kingma14}.
\texttt{PLAIN} means a plain MLP without batch normalization.
\texttt{BNA} and \texttt{BNB} are two different
implementations of BN, depending on whether BN is performed right before
(\texttt{BNB}) or right after (\texttt{BNA}) the activation of the hidden units.
They both use a re-scaling parameter to ensure enough flexibility of the
parametric structure~\citep{ioffe15}.  The epsilon parameter of both
\texttt{BNA} and \texttt{BNB} is set to $10^{-3}$.  For \texttt{RNGD}, we set
empirically $T=100$ and $\lambda=0.995$. The right table shows the mean and
standard deviation of the \emph{$\tau$-sharp ratio}, defined as the training
cost over the last $\tau{M}$ mini-batch iterations, where $M=55000$ is the total
number of iterations.

We can see that RNGD can significantly improve the learning curve over the other
methods and achieve a smaller sharp ratio.  The mechanism is similar to the first
experiment.  By \cref{eq:fimrelu}, $\nu_{\relu}(\bm{w}_i,\bm{x})$ is
approximately binary, emphasizing such \emph{informative samples} with
$\bm{w}_i^\intercal\tilde{\bm{x}}>0$, which are the ones contributing to the
learning of $\bm{w}_i$ with non-zero gradient values. Each output neuron has a
different subset of informative samples. While BN normalizes the layer input
$\bm{x}$ regardless of the output neurons, RNGD normalizes $\bm{x}$ differently
wrt different output, so that the informative samples for each output neuron is
centered and decorrelated.

As shown in \cref{fig:toy2}, \texttt{RNGD} appears to have a larger variation
during learning. This is because of the non-smooth update of the inverse metric.
One can however see from the right table that the variation is actually not
much, as the y-axis is in log-scale.

By the results of \texttt{RNGD}, it is clear that this MLP structure overfits
the input data. It is a fundamental trade-off between fitting the input data and
generalizing. This objective of this experiment is mainly focused on improving
learning optimization. The overfitting can be tackled by early stopping.

RNGD's computational time per each iteration is much more than the other
methods. In our experiments on a GPU instance of Amazon EC2, RNGD costs around
half a minute per each epoch, while the other methods only costs seconds. On
CPUs RNGD is even more time consuming. This is both due to the inefficiency of
our implementation and its algorithm complexity. To seek efficient RNGD
implementations is left for future work.

\section{Conclusion and discussion}\label{sec:con}

We propose to investigate local structures of large learning systems using the
new concept of Relative Fisher Information Matrix (RFIM).  The key advantage of this approach is that the local dynamics
can be analyzed in an accurate way without approximation. We present a core list of
 such local structures in neuron networks, and give their corresponding RFIMs. This
list of recipes can be used to provide guiding principles to neuron networks.
As a first example, we demonstrated how to apply single neuron RFIM and one
layer RFIM to improve the learning curve by using the relative natural gradient descent. 

Our work applies to mirror descent as well since natural gradient is related to mirror descent~\citep{mirror-ng-2015} as follows:
In mirror descent, given a strictly convex distance function $D(\cdot,\cdot)$  in the first argument (playing the role of the proximity function), we
express the gradient descent step as :
$$
\Theta_{t+1}=\arg\min_\Theta \{ \Theta^\top \nabla F(\Theta_t) + \frac{1}{\gamma} D(\Theta,\Theta_t)\}.
$$
When $D(\Theta,\Theta')$ is chosen as a Bregman divergence $B_F((\Theta,\Theta')=F(\Theta)-F(\Theta')-(\Theta-\Theta')^\top \nabla F(\Theta)$, it has been proved that
the mirror descent on the $\Theta$-parameterization is equivalent~\citep{mirror-ng-2015} to the natural gradient optimization on the induced Riemannian manifold with metric tensor ($\nabla^2 F(\Theta)$)
parameterized by the dual coordinate system $H=\nabla F(\Theta)$.
In general, to perform a Riemannian gradient descent for minimizing a real-valued function $f(\Theta)$ on the manifold, 
one needs to choose a proper metric tensor given in matrix form $G(\Theta)$. 
Thomas~\citeyearpar{GeNGA-2014} constructed a toy example showing that the natural gradient may diverge while the ordinary gradient (for $G=I$, the identity matrix) converges.
Recently, Thomas et al.~\citeyearpar{ENG-2016} proposed a new kind of descent method based on what they called the Energetic Natural Gradient that generalizes the natural gradient.
The energy distance $D_E(p(\Theta_1),p(\Theta_2))^2=E[2d_{p(\Theta_1)}(X,Y)-d_{p(\Theta_1)}(X,X')-d_{p(\Theta_1)}(Y,Y')]$ where $X,X'\sim p(\Theta_1)$
 and  $Y,Y'\sim p(\Theta_2)$, where $d_{p(\Theta_1)}(\cdot,\cdot)$ is a distance metric over the support.
Using a Taylor's expansion on their energy distance, they get the Energy Information Matrix (in a way similar to recovering the FIM from a Taylor's expansion of any $f$-divergence like the Kullback-Leibler divergence).
Their idea is to incorporate prior knowledge on the structure of the support (observation space) to define energy distance. 
Twisting the geometry of the support (say, Wassertein's optimal transport) with the geometry of the parametric distributions (Fisher-Rao geodesic distances) is indeed important~\citep{FROT-2015}.
In information geometry, invariance on the support is provided by a Markov morphism that is a probabilistic mapping of support to itself~\citep{cencov-1982}.
Markov morphism include deterministic transformation of a random variable by a statistic.
It is well-known that $I_{T}(\Theta)\prec I_X(\Theta)$ with equality iff. $T=T(X)$ is a sufficient statistic of $X$.
Thus to get the same invariance for the energy distance~\citep{ENG-2016}, one shall further require $d_{p(\Theta)}(T(X),T(Y))=d_{p(\Theta)}(X,Y)$.  

In the foreseeable future, we believe
that Relative Fisher Information Metrics (RFIMs) will provide a sound methodology to build further efficient learning
techniques in deep learning.

Our implementation is available at \url{https://www.lix.polytechnique.fr/~nielsen/RFIM}.

\bibliography{natural_gradient}

\appendix
\section{Non-linear Activation Functions}

By definition,
\begin{align}
\tanh(t)
&\defeq \frac{\exp(t) - \exp(-t)}
{\exp(t) + \exp(-t)},\label{eq:tanh}
\end{align}
and
\begin{align}
\sech(t)  &\defeq \frac{2}{\exp(t) + \exp(-t)}.
\end{align}
It is easy to verify that
\begin{equation}
\mathtt{sech}^2(t)
=
\left[ 1+\mathtt{tanh}(t) \right]
\left[ 1-\mathtt{tanh}(t) \right]
=
1 - \mathtt{tanh}^2(t).
\end{equation}

By \cref{eq:tanh},
\begin{align}
\mathtt{tanh}'(t)
&= \frac{\exp(t)+\exp(-t)}{\exp(t)+\exp(-t)}
-
\frac{ \exp(t) - \exp(-t) }{\left[ \exp(t) + \exp(-t) \right]^2}
\left[ \exp(t) - \exp(-t) \right]\nonumber\\
&=
\frac{ \left[ \exp(t) + \exp(-t) \right]^2 -
\left[ \exp(t) - \exp(-t) \right]^2 }{ \left[ \exp(t) + \exp(-t) \right]^2 }
=
\frac{4}{ \left[ \exp(t) + \exp(-t) \right]^2 }
=
\mathtt{sech}^2(t).
\end{align}

By definition, 
\begin{align}
\sigm(t) &\defeq \frac{1}{1 + \exp(-t)}.
\end{align}
Therefore
\begin{align}
\sigm'(t)
&= - \frac{1}{\left[1+\exp(-t)\right]^2}
\left(-\exp(-t)\right)
=
\frac{\exp(-t)}{\left[1+\exp(-t)\right]^2}
=
\sigm(t) \left[ 1-\sigm(t) \right].
\end{align}

By definition,
\begin{align}
\relu_\omega(t)
\defeq
\omega \ln\left(
\exp\left(\frac{\iota{t}}{\omega}\right)
+ \exp\left(\frac{t}{\omega}\right)
\right),
\end{align}
where $\omega>0$ and $0\le\iota<1$. Then,
\begin{align}\label{eq:drelu}
\relu'_\omega(t)
&=
\omega \frac{1}{\exp\left(\frac{\iota{t}}{\omega}\right) + \exp\left(\frac{t}{\omega}\right)}
\left(
\frac{\iota}{\omega}
\exp\left(\frac{\iota{t}}{\omega}\right)
+
\frac{1}{\omega}
\exp\left(\frac{t}{\omega}\right)
\right)\nonumber\\
&=
\frac
{\iota\exp\left(\frac{\iota{t}}{\omega}\right)
+ \exp\left(\frac{t}{\omega}\right)}
{\exp\left(\frac{\iota{t}}{\omega}\right) +
\exp\left(\frac{t}{\omega}\right)}\nonumber\\
&=
\iota
+(1-\iota)\frac{\exp\left(\frac{t}{\omega}\right)}
{\exp\left(\frac{\iota{t}}{\omega}\right) +
\exp\left(\frac{t}{\omega}\right)}\nonumber\\
&=
\iota
+(1-\iota)\frac{1}
{\exp\left((\iota-1)\frac{t}{\omega}\right) + 1}\nonumber\\
&=
\iota +(1-\iota)\sigm\left( \frac{1-\iota}{\omega} t \right).
\end{align}

By definition,
\begin{equation}
\elu(t)
=\left\{
\begin{array}{ll}
t & \text{if }t\ge0\\
\alpha \left( \exp(t) - 1 \right) & \text{if }t<0.\\
\end{array}
\right.
\end{equation}
Therefore
\begin{equation}\label{eq:eluprime}
\elu'(t)
=\left\{
\begin{array}{ll}
1              & \text{if }t\ge0\\
\alpha \exp(t) & \text{if }t<0.
\end{array}
\right.
\end{equation}

\section{A Single $\tanh$ Neuron}\label{sec:tanh}

Consider a neuron with parameters $\bm{w}$ and a Bernoulli output $y\in\{+,-\}$,
$p(y=+)=p^+$, $p(y=-)=p^-$, and $p^+ + p^-=1$. By the definition of RFIM, we have
\begin{align}
g^y(\bm{w}) &=
p^+
\frac{\partial\ln{p^+}}{\partial\bm{w}}
\frac{\partial\ln{p^+}}{\partial\bm{w}^\intercal}
+
p^-
\frac{\partial\ln{p^-}}{\partial\bm{w}}
\frac{\partial\ln{p^-}}{\partial\bm{w}^\intercal}
\nonumber\\
&=
\frac{1}{p^+}
\frac{\partial{p^+}}{\partial\bm{w}}
\frac{\partial{p^+}}{\partial\bm{w}^\intercal}
+
\frac{1}{p^-}
\frac{\partial{p^-}}{\partial\bm{w}}
\frac{\partial{p^-}}{\partial\bm{w}^\intercal}.
\end{align}
Since $p^++p^-=1$,
\begin{align}
\frac{\partial{p^+}}{\partial\bm{w}} + \frac{\partial{p^-}}{\partial\bm{w}} =0.
\end{align}
Therefore, the RFIM of a Bernoulli neuron has the general form
\begin{align}\label{eq:fimber}
g^y(\bm{w}) &=
\left( \frac{1}{p^+} + \frac{1}{p^-} \right)
\frac{\partial{p^+}}{\partial\bm{w}}
\frac{\partial{p^+}}{\partial\bm{w}^\intercal}
=
\frac{1}{p^+p^-}
\frac{\partial{p^+}}{\partial\bm{w}}
\frac{\partial{p^+}}{\partial\bm{w}^\intercal}.
\end{align}

A single $\mathtt{tanh}$ neuron with stochastic output $y\in\{-1,1\}$ is
given by
\begin{align}
p(y=-1)  &= \frac{1-\mu(\bm{x})}{2},\label{eq:y0}\\
p(y=1)   &= \frac{1+\mu(\bm{x})}{2},\label{eq:y1}\\
\mu(\bm{x}) &= \tanh( \bm{w}^\intercal \tilde{\bm{x}} ).\label{eq:mu}
\end{align}
By \cref{eq:fimber},
\begin{align}
g^y(\bm{w}) &=
\frac{1}{
\frac{1-\mu(\bm{x})}{2}
\frac{1+\mu(\bm{x})}{2}}
\left(
\frac{1}{2} \frac{\partial\mu}{\partial\bm{w}}
\right)
\left(
\frac{1}{2} \frac{\partial\mu}{\partial\bm{w}^\intercal}
\right)\nonumber\\
&=
\frac{1}{ \left( 1-\mu(\bm{x}) \right) \left( 1+\mu(\bm{x}) \right) }
\left[ 1-\mu^2(\bm{x}) \right]^2
\tilde{\bm{x}} \tilde{\bm{x}}^\intercal\nonumber\\
&=
\left[ 1-\mu^2(\bm{x}) \right]
\tilde{\bm{x}} \tilde{\bm{x}}^\intercal\nonumber\\
&=
\left[ 1 - \tanh^2( \bm{w}^\intercal \tilde{\bm{x}} ) \right]
\tilde{\bm{x}} \tilde{\bm{x}}^\intercal\nonumber\\
&=
\sech^2( \bm{w}^\intercal \tilde{\bm{x}} )
\tilde{\bm{x}} \tilde{\bm{x}}^\intercal.
\end{align}

An alternatively analysis is given as follows. By \cref{eq:y0,eq:y1,eq:mu},
\begin{align}
p(y=-1)
&= \frac{ \exp( -\bm{w}^\intercal \tilde{\bm{x}} ) }
{ \exp( \bm{w}^\intercal \bm{x} ) + \exp(-\bm{w}^\intercal \bm{x}) },\\
p(y=1) &= 
\frac{ \exp(\bm{w}^\intercal \tilde{\bm{x}}) }
{ \exp( \bm{w}^\intercal \tilde{\bm{x}} )
+ \exp(-\bm{w}^\intercal \tilde{\bm{x}} ) }.
\end{align}
Then,
\begin{align}
g^y(\bm{w})
&=
E_{y\sim{p(y\,\vert\,\bm{x})}}\left(
- \frac{\partial^2\ln{p}(y)}
{ \partial\bm{w} \partial\bm{w}^\intercal }
\right)\nonumber\\
&=
\frac{ \partial^2 }
{ \partial\bm{w} \partial\bm{w}^\intercal }
\ln \left[ \exp(  \bm{w}^\intercal \tilde{\bm{x}} )
         + \exp( -\bm{w}^\intercal \tilde{\bm{x}} ) \right]
\quad\text{(first linear term vanishes)}
\nonumber\\
&=
\frac{\partial}{\partial\bm{w}^\intercal}
\left[
\frac{ \exp(\bm{w}^\intercal \tilde{\bm{x}} ) - \exp( -\bm{w}^\intercal \tilde{\bm{x}} ) }
     { \exp(\bm{w}^\intercal \tilde{\bm{x}} ) + \exp( -\bm{w}^\intercal \tilde{\bm{x}} ) }
\right] \tilde{\bm{x}}
\nonumber\\
&=
\frac{\partial}{\partial\bm{w}^\intercal}
\tanh(\bm{w}^\intercal \tilde{\bm{x}}) \tilde{\bm{x}}
\nonumber\\
&=
\sech^2(\bm{w}^\intercal \tilde{\bm{x}})
\tilde{\bm{x}} \tilde{\bm{x}}^\intercal.
\end{align}

The intuitive meaning of $g^y(\bm{w})$ is a weighted covariance to emphasize
such ``informative'' $\bm{x}$'s that 
\begin{itemize}
\item are in the linear region of $\tanh$
\item contain ``ambiguous'' samples
\end{itemize}
We will need at least $\dim(\bm{w})$ samples to make $g^y(\bm{w})$ full rank.

\section{A Single $\sigm$ Neuron}
A single $\sigm$ neuron is given by
\begin{align}
p(y=0) &= 1-\mu(\bm{x}),\\
p(y=1) &= \mu(\bm{x}),\\
\mu(\bm{x}) &= \sigm( \bm{w}^\intercal \tilde{\bm{x}} ).
\end{align}

By \cref{eq:fimber},
\begin{align}
g^y(\bm{w}) &= 
\frac{1}{ p(y=0)p(y=1) }
\frac{\partial{p}(y=1)}{\partial\bm{w}}
\frac{\partial{p}(y=1)}{\partial\bm{w}^\intercal}\nonumber\\
&=
\frac{1}{\mu(\bm{x}) (1-\mu(\bm{x}))}
\frac{\partial\mu}{\partial\bm{w}}
\frac{\partial\mu}{\partial\bm{w}^\intercal}\nonumber\\
&=
\frac{1}{\mu(\bm{x}) (1-\mu(\bm{x}))}
\mu^2(\bm{x}) (1-\mu(\bm{x}))^2
\tilde{\bm{x}} \tilde{\bm{x}}^\intercal\nonumber\\
&=
\mu(\bm{x}) (1-\mu(\bm{x}))
\tilde{\bm{x}} \tilde{\bm{x}}^\intercal\nonumber\\
&=
\sigm( \bm{w}^\intercal \tilde{\bm{x}} )
\left[ 1 - \sigm( \bm{w}^\intercal \tilde{\bm{x}} ) \right]
\tilde{\bm{x}} \tilde{\bm{x}}^\intercal.
\end{align}

\section{A Single $\relu$ Neuron}\label{sec:relu}

Consider a single neuron with Gaussian output $p(y\,\vert\,\bm{w},\bm{x})
=G(y\,\vert\,\mu(\bm{w},\bm{x}), \sigma^2)$. Then
\begin{align}\label{eq:fimgauss}
g^y(\bm{w}\,\vert\,\bm{x})
&= E_{p(y\,\vert\,\bm{w},\bm{x})}\left[ 
\frac{\partial\ln{G}(y\,\vert\,\mu,\sigma^2)}{\partial\bm{w}}
\frac{\partial\ln{G}(y\,\vert\,\mu,\sigma^2)}{\partial\bm{w}^\intercal}
\right]\nonumber\\
&= E_{p(y\,\vert\,\bm{w},\bm{x})}\left[
\frac{\partial}{\partial\bm{w}}
\left(-\frac{1}{2\sigma^2} ( y - \mu )^2\right)
\frac{\partial}{\partial\bm{w}^\intercal}
\left(-\frac{1}{2\sigma^2} ( y - \mu )^2\right)
\right]\nonumber\\
&= E_{p(y\,\vert\,\bm{w},\bm{x})}\left[
\left(- \frac{1}{\sigma^2} (\mu-y) \right)^2
\frac{\partial\mu}{\partial\bm{w}}
\frac{\partial\mu}{\partial\bm{w}^\intercal}
\right]\nonumber\\
&=\frac{1}{\sigma^4}
E_{p(y\,\vert\,\bm{w},\bm{x})} \left( \mu - y \right)^2
\frac{\partial\mu}{\partial\bm{w}}
\frac{\partial\mu}{\partial\bm{w}^\intercal}\nonumber\\
&=
\frac{1}{\sigma^2}
\frac{\partial\mu}{\partial\bm{w}}
\frac{\partial\mu}{\partial\bm{w}^\intercal}.
\end{align}

A single $\relu$ neuron is given by
\begin{equation}
\mu(\bm{w},\bm{x}) = \relu_\omega( \bm{w}^\intercal\tilde{\bm{x}} ).
\end{equation}
By \cref{eq:fimgauss,eq:drelu},
\begin{align}\label{eq:fimreluapp}
g^y(\bm{w})
&=\frac{1}{\sigma^2}
\left[
\iota +(1-\iota)\sigm\left( \frac{1-\iota}{\omega} \bm{w}^\intercal\tilde{\bm{x}} \right)
\right]^2
\tilde{\bm{x}} \tilde{\bm{x}}^\intercal.
\end{align}

\section{A Single $\elu$ Neuron}\label{sec:elu}

Similar to the analysis in \cref{sec:relu}, a single $\elu$ neuron is given by
\begin{equation}
\mu(\bm{w},\bm{x}) = \elu(\bm{w}^\intercal\tilde{\bm{x}}).
\end{equation}
By \cref{eq:eluprime},
\begin{equation}
\frac{\partial\mu}{\partial\bm{w}}
=
\left\{
\begin{array}{ll}
\tilde{\bm{x}}
& \text{if }\bm{w}^\intercal\tilde{\bm{x}}\ge0\\
\alpha \exp( \bm{w}^\intercal\tilde{\bm{x}} ) 
\tilde{\bm{x}}
& \text{if }\bm{w}^\intercal\tilde{\bm{x}}<0.
\end{array}
\right.
\end{equation}
By \cref{eq:fimgauss},
\begin{equation}
g^y(\bm{w})
= \left\{
\begin{array}{ll}
\frac{1}{\sigma^2} \tilde{\bm{x}}\tilde{\bm{x}}^\intercal
& \text{if }\bm{w}^\intercal\tilde{\bm{x}}\ge0\\
\frac{1}{\sigma^2} \left( \alpha \exp(
\bm{w}^\intercal\tilde{\bm{x}}
) \right)^2
\tilde{\bm{x}}\tilde{\bm{x}}^\intercal
& \text{if }\bm{w}^\intercal\tilde{\bm{x}}<0.\\
\end{array}
\right.
\end{equation}

\section{RFIM of a Linear Layer}
Consider a linear layer
\begin{equation}
p(\bm{y})
=
G\left( \bm{y}\,\vert\,\bm{W}^\intercal\tilde{\bm{x}}, \sigma^2\bm{I} \right),
\end{equation}
where $\bm{W}=(\bm{w}_1,\cdots,\bm{w}_{D_y})$. By the definition of multivariate
Gaussian distribution,
\begin{equation}
\ln p(\bm{y})
=
-\frac{1}{2}\ln{2\pi}
-\frac{D_y}{2} \ln \sigma^2
-\frac{1}{2\sigma^2} \sum_{i=1}^{D_y} \left( y_i - \bm{w}_i^\intercal \tilde{\bm{x}} \right)^2.
\end{equation}
Therefore,
\begin{equation}
\forall{i},\quad
\frac{\partial}{\partial\bm{w}_i} \ln p(\bm{y}) = 
-\frac{1}{\sigma^2} \left( \bm{w}_i^\intercal \tilde{\bm{x}} - y_i \right) \tilde{\bm{x}}.
\end{equation}
Therefore,
\begin{equation}
\forall{i},\forall{j}\quad
\frac{\partial}{\partial\bm{w}_i} \ln p(\bm{y})
\frac{\partial}{\partial\bm{w}_j^\intercal} \ln p(\bm{y})
= 
\frac{1}{\sigma^4}
\left( y_i - \bm{w}_i^\intercal \tilde{\bm{x}} \right)
\left( y_j - \bm{w}_j^\intercal \tilde{\bm{x}} \right)
\tilde{\bm{x}}
\tilde{\bm{x}}^\intercal.
\end{equation}
$\bm{W}$ is vectorized by stacking its columns $\{\bm{w}_i\}_{i=1}^{D_y}$.
In the following $\bm{W}$ will be used interchangeably to denote either
the matrix or its vector form.
Correspondingly, the RFIM $g^{\bm{y}}(\bm{W})$ has $D_y\times{D_y}$ blocks,
where the off-diagonal blocks are
\begin{equation}
\forall{i\neq{j}},\quad
E_{p(\bm{y})}
\left(
\frac{\partial}{\partial\bm{w}_i} \ln p(\bm{y})
\frac{\partial}{\partial\bm{w}_j^\intercal} \ln p(\bm{y})
\right)
=
\frac{1}{\sigma^4}
E_{p(\bm{y})} \left[
\left( y_i - \bm{w}_i^\intercal \tilde{\bm{x}} \right)
\left( y_j - \bm{w}_j^\intercal \tilde{\bm{x}} \right) \right]
\tilde{\bm{x}}
\tilde{\bm{x}}^\intercal
=
\bm0,
\end{equation}
and the diagonal blocks are 
\begin{equation}
\forall{i},\quad
E_{p(\bm{y})}
\left(
\frac{\partial}{\partial\bm{w}_i} \ln p(\bm{y})
\frac{\partial}{\partial\bm{w}_i^\intercal} \ln p(\bm{y})
\right)
=
\frac{1}{\sigma^4}
E_{p(\bm{y})} \left( y_i - \bm{w}_i^\intercal \tilde{\bm{x}} \right)^2
\tilde{\bm{x}} \tilde{\bm{x}}^\intercal
=
\frac{1}{\sigma^2} \tilde{\bm{x}} \tilde{\bm{x}}^\intercal.
\end{equation}
In summary, 
\begin{equation}
g^{\bm{y}}(\bm{W})
=
\frac{1}{\sigma^2} \diag\left[
\tilde{\bm{x}} \tilde{\bm{x}}^\intercal,
\cdots,
\tilde{\bm{x}} \tilde{\bm{x}}^\intercal \right].
\end{equation}

\section{RFIM of a Non-Linear Layer}

The statistical model of a non-linear layer is
\begin{equation}
p(\bm{y}\,\vert\,\bm{W}, \bm{x})
= \prod_{i=1}^{D_y} p(y_i\,\vert\,\bm{w}_i, \bm{x}).
\end{equation}
Then,
\begin{equation}
\ln{p}(\bm{y}\,\vert\,\bm{W}, \bm{x})
= \sum_{i=1}^{D_y} \ln{p}(y_i\,\vert\,\bm{w}_i, \bm{x}).
\end{equation}
Therefore,
\begin{equation}
\frac{\partial^2}{\partial\bm{W}\partial\bm{W}^\intercal}
\ln{p}(\bm{y}\,\vert\,\bm{W}, \bm{x})
=
\begin{bmatrix}
\frac{\partial^2}{\partial\bm{w}_1\partial\bm{w}_1^\intercal}
\ln{p}(y_1\,\vert\,\bm{w}_1, \bm{x}) && \\
& \ddots & \\
&& \frac{\partial^2}
{\partial\bm{w}_{D_y}\partial\bm{w}_{D_y}^\intercal}
\ln{p}(y_{D_y}\,\vert\,\bm{w}_{D_y}, \bm{x})\\
\end{bmatrix}.
\end{equation}
Therefore RFIM $g^{\bm{y}}(\bm{W})$ is a block-diagonal matrix, with the $i$'th
block given by 
\begin{equation}
-E_{{p}(\bm{y}\,\vert\,\bm{W}, \bm{x})}
\left[
\frac{\partial^2}
{\partial\bm{w}_i\partial\bm{w}_i^\intercal}
\ln{p}(y_i\,\vert\,\bm{w}_i, \bm{x})
\right]
=
-E_{{p}(y_i\,\vert\,\bm{w}_i, \bm{x})}
\left[
\frac{\partial^2}
{\partial\bm{w}_i\partial\bm{w}_i^\intercal}
\ln{p}(y_i\,\vert\,\bm{w}_i, \bm{x})
\right],
\end{equation}
which is simply the single neuron RFIM of the $i$'th neuron.

\section{RFIM of a Softmax Layer}

Recall that
\begin{equation}
\forall{i}\in\left\{ 1,\cdots,m \right\},\quad
p(y=i) =
\frac{\exp(\bm{w}_i\tilde{\bm{x}})}
{ \sum_{i=1}^m \exp(\bm{w}_i\tilde{\bm{x}}) }.
\end{equation}
Then 
\begin{equation}
\forall{i},\quad
\ln{p}(y=i) =
\bm{w}_i\tilde{\bm{x}} - \ln\sum_{i=1}^m \exp(\bm{w}_i\tilde{\bm{x}}).
\end{equation}
Hence
\begin{equation}
\forall{i},~\forall{j},\quad
\frac{\partial\ln{p}(y=i)} {\partial\bm{w}_j}
=
\delta_{ij}\tilde{\bm{x}} -
\frac{ \exp(\bm{w}_j\tilde{\bm{x}}) }
{\sum_{i=1}^m \exp(\bm{w}_i\tilde{\bm{x}})}
\tilde{\bm{x}},
\end{equation}
where $\delta_{ij}=1$ if and only if $i=j$ and $\delta_{ij}=0$ otherwise. Then
\begin{align}\label{eq:softmaxderi}
\forall{i},~\forall{j},~\forall{k},\quad
\frac{\partial^2\ln{p}(y=i)} {\partial\bm{w}_j \partial\bm{w}_k^\intercal}
&=
- \delta_{jk}
\frac{ \exp(\bm{w}_j\tilde{\bm{x}}) }
{\sum_{i=1}^m \exp(\bm{w}_i\tilde{\bm{x}})}
\tilde{\bm{x}} \tilde{\bm{x}}^\intercal
+
\frac{ \exp(\bm{w}_j\tilde{\bm{x}}) }
{\left(\sum_{i=1}^m \exp(\bm{w}_i\tilde{\bm{x}})\right)^2}
\exp(\bm{w}_k\tilde{\bm{x}})
\tilde{\bm{x}} \tilde{\bm{x}}^\intercal
\nonumber\\
&=
\left( -\delta_{jk} \eta_j + \eta_j \eta_k \right)
\tilde{\bm{x}} \tilde{\bm{x}}^\intercal.
\end{align}
The right-hand-side of \cref{eq:softmaxderi} does not depend on $i$. Therefore
\begin{equation}
g^{\bm{y}}(\bm{W})
=
\begin{bmatrix}
(\eta_1 - \eta_1^2) \tilde{\bm{x}} \tilde{\bm{x}}^\intercal & 
-\eta_1\eta_2  \tilde{\bm{x}} \tilde{\bm{x}}^\intercal & 
\cdots & - \eta_1\eta_m \tilde{\bm{x}} \tilde{\bm{x}}^\intercal \\
-\eta_2\eta_1  \tilde{\bm{x}} \tilde{\bm{x}}^\intercal & 
(\eta_2 - \eta_2^2) \tilde{\bm{x}} \tilde{\bm{x}}^\intercal 
&\cdots&
- \eta_2\eta_m \tilde{\bm{x}} \tilde{\bm{x}}^\intercal \\
\vdots & \vdots &\ddots & \vdots \\
-\eta_m\eta_1 \tilde{\bm{x}} \tilde{\bm{x}}^\intercal &
-\eta_m\eta_2 \tilde{\bm{x}} \tilde{\bm{x}}^\intercal &
\cdots &
(\eta_m - \eta_m^2) \tilde{\bm{x}} \tilde{\bm{x}}^\intercal \\
\end{bmatrix}.
\end{equation}

\section{RFIM of Two layers}

Consider a two layer structure, where the output $\bm{y}$ satisfies a
multivariate Bernoulli distribution with independent dimensions.
By a similar analysis to \cref{sec:tanh}, we have
\begin{align}\label{eq:divtwolayer}
g^{\bm{y}}(\bm{W})
&=
\sum_{l=1}^{D_y}
\nu_{f}(\bm{c}_l,\bm{h})
\frac{\partial{\bm{c}_l^\intercal\bm{h}}}{\partial\bm{W}}
\frac{\partial{\bm{c}_l^\intercal\bm{h}}}{\partial\bm{W}^\intercal}.
\end{align}
It can be written block by block as 
$g^{\bm{y}}(\bm{W}) = \left[ \bm{G}_{ij} \right]_{D_h\times{}D_h}$, where
each block $\bm{G}_{ij}$ means the correlation between the $i$'th hidden
neuron with weights $\bm{w}_i$ and the $j$'th hidden neuron with
weights $\bm{w}_j$. By \cref{eq:divtwolayer},
\begin{align}
\bm{G}_{ij} 
&= 
\sum_{l=1}^{D_y}
\nu_{f}(\bm{c}_l,\bm{h})
\frac{\partial{\bm{c}_l^\intercal\bm{h}}}{\partial\bm{w}_i}
\frac{\partial{\bm{c}_l^\intercal\bm{h}}}{\partial\bm{w}_j^\intercal}
=
\sum_{l=1}^{D_y}
\nu_{f}(\bm{c}_l,\bm{h})
\frac{\partial{c_{il}h_i}}{\partial\bm{w}_i}
\frac{\partial{c_{jl}h_j}}{\partial\bm{w}_j^\intercal}\nonumber\\
&=
\sum_{l=1}^{D_y}
\nu_{f}(\bm{c}_l,\bm{h}) c_{il}c_{jl}
\frac{\partial{h_i}}{\partial\bm{w}_i}
\frac{\partial{h_j}}{\partial\bm{w}_j^\intercal}
=
\sum_{l=1}^{D_y}
\nu_{f}(\bm{c}_l, \bm{h}) c_{il}c_{jl}
\left( \nu_{f}(\bm{w}_i, \bm{x}) \tilde{\bm{x}} \right)
\left( \nu_{f}(\bm{w}_j, \bm{x}) \tilde{\bm{x}}^\intercal \right)\nonumber\\
&=
\sum_{l=1}^{D_y}
c_{il}c_{jl}
\nu_{f}(\bm{c}_l, \bm{h})
\nu_{f}(\bm{w}_i, \bm{x})
\nu_{f}(\bm{w}_j, \bm{x})
\tilde{\bm{x}}\tilde{\bm{x}}^\intercal.
\end{align}

The proof of the other case, where two $\relu$ layers have stochastic output
$\bm{y}$ satisfying a multivariate Gaussian distribution with independent
dimensions, is very similar and is omitted.

\end{document}